\documentclass[preprint,10pt,authoryear]{elsarticle}
\usepackage[margin=1in]{geometry}
\usepackage{hyperref}
\usepackage[utf8]{inputenc} 
\usepackage[T1]{fontenc}    
\usepackage{url}            
\usepackage{booktabs}       
\usepackage{amsfonts}       
\usepackage{nicefrac}       
\usepackage{microtype}      
\usepackage{lipsum}
\usepackage{amsmath}
\usepackage{amssymb}
\usepackage{amsthm}
\usepackage{bm}
\usepackage{url}
\usepackage{stackengine}
\usepackage{caption}
\usepackage{makecell}
\usepackage{multirow}
\usepackage{mwe}
\usepackage{subcaption}
\usepackage{graphicx}
\usepackage{diagbox}
\usepackage{soul,color}
\usepackage[user,titleref]{zref}
\usepackage[ruled,vlined]{algorithm2e}
\usepackage{lineno}
\journal{Information Fusion}

\begin{document}
\begin{frontmatter}






\title{Multi-Class Traffic Assignment using Multi-View Heterogeneous Graph Attention Networks}

\author[inst1]{Tong Liu\footnote{Corresponding Author, tongl5@illinois.edu}}
\author[inst1]{Hadi Meidani}
\affiliation[inst1]{organization={University of Illinois, Urbana-Champaign, Department of Civil and Environmental Engineering},
            addressline={205 N Mathews Ave}, 
            city={Urbana},
            postcode={61801}, 
            state={IL},
            country={USA}}

\begin{abstract}
Solving traffic assignment problem for large networks is computationally challenging when conventional optimization-based methods are used. In our research, we  develop an innovative surrogate model for a traffic assignment when   multi-class vehicles are involved. We do so by employing heterogeneous graph neural networks which uses a multiple-view graph attention mechanism tailored to different vehicle classes, along with additional links connecting origin-destination pairs. We also integrate the node-based flow conservation law into the loss function. As a result, our model adheres to flow conservation while delivering highly accurate predictions for link flows and utilization ratios. Through numerical experiments conducted on urban transportation networks, we demonstrate that our model surpasses traditional neural network approaches in convergence speed and predictive accuracy in both user equilibrium and system optimal versions of traffic assignment. 
\end{abstract}


\end{frontmatter}


\section{Introduction}
\label{sec:introduction}

Traffic assignment is crucial for analyzing transportation networks, as it enhances our understanding of traffic flow dynamics and provides insights into congestion and environmental impacts \citep{nie2004models}. The primary goal of addressing the traffic assignment problem (TAP) is to analyze traffic flow patterns and identify congestion bottlenecks within road networks. This analysis aids urban planners in devising strategic efforts to alleviate traffic congestion \citep{seliverstov2017development}. Beyond network analysis, traffic assignment has significant implications for regional resilience and urban development. Urban planners utilize traffic assignment to assess network robustness during severe events such as hurricanes \citep{zou2020resilience} and earthquakes \citep{liu2023optimizing}. These assessments help ensure that transportation infrastructure meets the city's evolving demands and standards.

There are two major principal approaches to resolving traffic assignment issues: user equilibrium (UE) assignment and system optimum (SO) assignment. Each method is grounded in distinct principles and is suited to different contexts. Traditionally, TAP has been addressed through optimization formulations, leading to the development of numerous optimization-based methods \citep{lee2003conjugate, zhang2023bi}. Recently, neural network techniques such as convolutional neural networks (CNNs) \citep{fan2023deep} and graph neural networks (GNNs) \citep{rahman2023data} have increasingly been applied to tackle these problems. Particularly when dealing with incomplete origin-destination (OD) demand data, GNNs often outperform optimization-based methods in terms of speed and accuracy, with encoder-decoder models demonstrating superior performance \citep{liu2024end}.

For instance, although scenarios with reduced link capacity are considered, situations where roads are completely closed due to regulatory orders, resulting in different network topology,  are often overlooked. Moreover, previous studies typically assume the presence of only a single vehicle class within the network, which does not accurately reflect real-world conditions. In practice, various vehicle classes exist, each with different levels of road accessibility. For example, sedans and small vehicles can navigate most highways, primary, and secondary roads, whereas trucks and other heavy-duty vehicles are generally confined to highways and primary roads, often subject to time-based restrictions. These critical factors have not been accounted for in previous research, thereby limiting the practical applicability and feasibility of the existing methods. These gaps highlight the  need for further research to enhance the adaptability and real-world relevance of neural networks in traffic management.

From the perspective of graph neural network formulation, the traffic flow patterns belonging to  each vehicle class can be conceptualized as a distinct ``view" within the transportation network. That is, each class of vehicle is related to its unique OD demands and available routing options, leading to distinct traffic patterns. This is while the route choice of vehicles of a certain class does in fact influence the routes of  other classes, creating interdependencies across these views. As a result, these different views, representing various vehicle classes, interact at the link level in the GNN, causing the interdependent views to shape the overall traffic flow pattern at a system-wide scale. Therefore, it is crucial to enable exchange of information (i.e. message passing) between different views within a GNN model for a multi-class vehicle traffic assignment. To the best of our knowledge, the proposed approach in this paper is the first work addressing this  research gap.

To model the interdependencies between the classes, we build upon the model we proposed in \citep{liu2024end}, which was a heterogeneous GNN consisting of physical roads and virtual "Origin-destination" links. We then introduce a novel multi-view heterogeneous graph attention network. Our approach uniquely assigns a distinct GNN view to each vehicle class, with each view utilizing specific features. This enhancement incorporates OD demand data from various vehicle classes, enabling more effective feature propagation across the network. Moreover, we propose a multi-view graph encoder-decoder structure that can be integrated with any existing GNN to calculate the solution of a traffic assignment problem. The encoder-decoder architecture is designed to capture interactions among different vehicle classes, facilitating the estimation of both link flows and link utilization ratios. Additionally, we incorporate a conservation-based loss function that promotes the flow predictions to align with established traffic flow principles. In summary, the key contributions of our study are threefold:
\begin{itemize}
    \item This is the first multi-view  GNN framework that can handle multi-class vehicle traffic assignment, effectively capturing the complex interdependencies between various vehicle classes and origin-destination demand.
    \item The model leverages the conservation of traffic flow principle to enhance the accuracy of flow predictions. 
    \item The proposed framework is demonstrably capable to predict traffic flows under extreme conditions, such as road closures, restrictions, and altered network topologies that specific to each vehicle classes, thereby broadening its utility in real-world transportation problems.
\end{itemize}

In this work we demonstrate the efficacy and generalizability of the proposed model through extensive experiments on various road network topologies, link properties, and OD demands. The remainder of this article is structured as follows. Section \ref{sec:literature} provides a review of related literature. General backgrounds on the traffic assignment problem, and the graph neural network  are presented in Section \ref{sec:background}. Section \ref{sec:architecture} includes the introduction of the proposed multi-view heterogeneous graph attention network for multi-class vehicle traffic assignment. Furthermore, the experiments with urban road networks are presented to demonstrate the accuracy and generalization capability of the proposed framework in Section \ref{sec:experiment}. Finally, the conclusion and discussion of the proposed framework are presented in Section \ref{sec:conslusion}.

\section{Related Literature}
\label{sec:literature}

Solving the traffic assignment problem (TAP) has been the focus of many studies. For example, \cite{lee2003conjugate} aimed to improve both convergence and computational efficiency in solving TAP for large-scale traffic networks by proposing a conjugate gradient projection method to enhance traditional gradient projection techniques. Similarly, \cite{babazadeh2020reduced} introduced a reduced gradient algorithm that selects non-basic paths to effectively manage computational complexity. In the context of multi-class vehicle traffic assignment, \cite{sun2021multi} proposed a logit-based multi-class ridesharing user equilibrium framework, formulated as a mixed complementarity problem. Their findings demonstrate that policies such as car restrictions, cordon tolling, and subsidization are significantly influenced by the relative performance of transit systems and users' preferences for driving. Additionally, \cite{xu2024range} addressed the range-constrained traffic assignment problem by considering heterogeneous range anxiety among electric vehicle drivers, showing that heightened anxiety can lead to severe congestion on critical network links.

Recently, neural networks have been successfully used in data reconstruction and generalization \citep{zhang2018missing}, making them a promising solution for solving TAPs while addressing the issue of missing origin-destination (OD) demands. However, the application of neural networks in TAPs remains relatively underexplored. A  prominent neural network approach in this domain is the use of convolutional neural networks (CNNs), where the transportation network is modeled as a grid map on which the incomplete OD demand is handled, as done in \cite{fan2023deep}. Another choice for the neural network architecture is the graph neural network (GNN). GNNs solve TAPs by capturing spatial information using  graph topology information \citep{liu2024graph}. For instance, \cite{rahman2023data} applied graph convolutional neural network to learn traffic flow patterns directly from OD demand and link flow data, offering a scalable solution for large-scale TAPs. Building upon this work, \citep{liu2024end} and \cite{liu2024heterogeneous} extended the homogeneous GNN into a heterogeneous GNN that utilizes two different types of links, namely roadway and OD links,  to estimate traffic flows in both static and dynamic traffic assignments. However, these approaches have primarily focused on single-class vehicle traffic assignment, and the interactions and interdependencies between different classes of vehicles are not considered. Furthermore,  most of the existing  neural network approaches  focus on in-distribution data prediction, and the traffic flow prediction under unseen scenarios or out-of-distribution data is not fully investigated.

In summary, despite significant advances enabled by neural networks, a  fast solution approach for multi-class vehicle traffic assignment and link-wise flow estimation is needed for enhanced traffic management for such problems. Specifically,  there is a need to address the interaction between different vehicle classes and to understand how multiple classes of vehicle collectively influence the network flow distribution.

\section{Technical Background}
\label{sec:background}

\subsection{Multi-Class Vehicle Traffic Assignment Problem}
The traffic assignment problem involves estimating  traffic volumes or flows on each edge in the network, given a graph topology and  OD demand values. Specifically, let us consider a transportation network represented as a graph $\mathcal{G} = (\mathcal{V}, \mathcal{E})$, where  $\mathcal{V}$ denotes the set of vertices (or nodes), which are the intersections of roads, and  $\mathcal{E}$ denotes the edges or roadways  connecting the nodes. Furthermore, consider $\mathcal{C}$ to be the set of vehicle classes. The general form of multi-class vehicle traffic assignment problem can then be formulated as an optimization optimization problem given by


\begin{equation}
    \label{eq:basic_formulation}
    \min_{f}: \quad \sum_{e \in \mathcal{E}} z_e \left(\sum_{c \in \mathcal{C}} f_{e,c}\right),
\end{equation}
where $f_{e,c}$ the flow  for vehicle class $c$ on link $e$, and  $z_e(\cdot)$ is the cost function for link $e$. The link cost function, which is a function of link flow, can represent the travel time, travel distance, or other relevant factors. Additionally, the traffic assignment problem can have different formulations depending on the specific objectives and assumptions. For instance, a multi-class vehicle user equilibrium traffic assignment problem, adapted from \citep{beckmann1956studies}, can be formulated as follows:

\begin{equation}
\label{eq:UE-TAP}
\begin{aligned}
    \min \quad &\sum_{e \in \mathcal{E}}  \int_0^{\sum_{c \in \mathcal{C}}y_{e,c}} t_{e}(\omega) \mathrm{d}\omega \\
    \mathrm{s.t.}\quad & \sum_k f_{k,c}^{rs} = q_{rs,c}, \ \quad \quad \quad \quad \forall r, s \in \mathcal{V}, \forall c \in \mathcal{C}, \\
    & f_{k,c}^{rs} \geq 0,\ \quad \quad \quad \quad \quad  \ \quad \quad \forall k, r, s \in \mathcal{V},  \forall c \in \mathcal{C}, \\
    & y_{e,c} = \sum_{rs} \sum_{k} f_{k,c}^{rs} \zeta^{rs}_{e,k,c}, \quad \forall e \in \mathcal{E}, \forall c \in \mathcal{C},
\end{aligned}
\end{equation}
where the objective function is the summation over all road segments of the integral of the link travel time function between 0 and the link flow, $t_e(\cdot)$ is the link travel time function, $q_{rs, c}$ is the total demand from source $r$ to destination $s$ for vehicle class $c$, $f^{rs}_k$ represents the flow of vehicle class $c$ on $k^{\mathrm{th}}$ path from $r$ to {s}. $\zeta^{rs}_{e,k, c}$ is the binary variable which equals 1 when link $e$ is on $k^{\mathrm{th}}$ path to connect $r$ and $s$ for vehicle class $c$. It is noted that the objective function in Beckmann's formulation serves more as a mathematical construct for optimization than a direct physical representation.

\subsection{Graph Neural Network}
\label{sec:gnn}
In the early architecture of fully connected neural networks applied to scientific problems, the input data structures are normally of a fixed shape, and the predominant applications processed Euclidean data \cite{liu2020automatic}. However, a non-Euclidean data structure such as graph-structured data is pervasive in engineering applications. The complexity and variability of the graph structure data make it inefficient to be processed with conventional neural network architectures. To address this challenge, GNNs are specifically designed to handle graph-structured data. GNNs operate on the node features and edge features and learn to extract embeddings from nodes and edges, aiming to capture the underlying patterns affecting the graph data.


GNNs consist of two major components. The first component is the graph topology, typically represented as a graph $\mathcal{G} = (\mathcal{V}, \mathcal{E})$, where $\mathcal{V}$ and $\mathcal{E}$ represent the node and edge set, respectively. The second component is the feature embedding, including initial node embedding $\bm{x}^{(0)}_{u} \in \mathbb{R}^{1 \times N_u}$ for $u \in \mathcal{V}$ and initial edge embedding $\bm{y}^{(k)}_{e} \in \mathbb{R}^{1 \times N_e}$ for $e \in \mathcal{E}$ in lower dimensional space, where $N_u$ and $N_e$ is the dimension of the feature embedding. 

GNNs typically involve an update rule at the node level. This update rule in turn involves an aggregation of information from neighboring nodes. There are various neighborhood aggregation approaches that can be effectively used in large-scale graphs. A prominent method is GraphSAGE \citep{hamilton2017inductive}, which diverges from spectral techniques by performing convolutions directly in the spatial domain. GraphSAGE operates by sampling a fixed-size neighborhood of nodes and aggregating their features to update the target node's representation. The update rule in GraphSAGE is defined by:

\begin{equation}
\label{eq:graphSAGE}
\begin{split}
    \bm{x}_{\mathcal{N}(i)}^{(k-1)} &= \mathrm{AGGREGATE}_k \left(\left[\bm{x}_j^{(k-1)} : j \in \mathcal{N}(i)\right]\right), \\
    \bm{x}_i^{(k)} & = \sigma \left(\bm{W}^{(k)} \cdot \left[\bm{x}_i^{(k-1)} | \bm{x}_{\mathcal{N}(i)}^{(k-1)} \right]\right), 
\end{split}
\end{equation} 
where $\bm{x}_i^{(k)}$ is the node embedding of node $i$ at the $k^{\mathrm{th}}$ layer, $\mathcal{N}(i)$ denotes the neighborhood of node $i$, $\mathrm{AGGREGATE}_k$ is a differentiable function such as mean, sum, or max pooling, $\sigma$ is a non-linear activation function, and $|$ denotes concatenation. The learnable parameters are represented by $\bm{W}^{(k)}$ at each layer $k$. This aggregation-based method allows for learning node embeddings that incorporate both local graph structure and node feature information, enabling the application of GraphSAGE to inductive learning tasks on graphs that exhibit unseen nodes during training. 

As a different approach to modeling graph data, the graph attention network (GAT) offers an innovative method by determining attention weights for each node, which are influenced by its own features and those of its adjacent nodes \citep{velivckovic2017graph}. In this approach, the updated representation of each node $i$ at the $k^{\mathrm{th}}$ layer, is computed using

\begin{equation}
\label{eq:gat_score}
    \bm{x}^{(k)}_i = \sigma\left(\sum_{j \in \mathcal{N}(i)} \alpha_{ij} \bm{W}_x \bm{x}^{(k-1)}_j\right),
\end{equation}
where $\bm{W}_x$ is a learnable weight matrix,  $\alpha_{ij}$ is the attention weight assigned to the node $j$ related to its neighboring node $i$, and $\sigma$ is an activation function. The attention weights are computed as follows:

\begin{equation}
\label{eq:gat_weight}
    \alpha_{ij} = \frac{\exp(\sigma(\bm{a}^T[\bm{W}_x \bm{x}^{(k-1)}_i | \bm{W}_x \bm{x}^{(k-1)}_j]))}{\sum_{v\in \mathcal{N}(i)} \exp(\sigma(\bm{a}^T[\bm{W}_x \bm{x}^{(k-1)}_i | \bm{W}_x \bm{x}^{(k-1)}_v]))}, 
\end{equation}
where $\bm{a}$ is a learnable vector of weights. By employing a multilayered attention mechanism, the network is capable of capturing progressively intricate graph representations. This adaptive focus on various nodes within their localities enhances the efficacy of the model.

The aforementioned formulation is valid for homogeneous graphs, where all nodes and edges have the same semantic meaning. However, it is noted that real-world graphs are not always homogeneous. For instance, in the literature citation graph, nodes can represent various entities such as papers, authors, and journals, while edges may denote different semantic relationships. When the graph contains different classes of nodes or edges, it is considered as a heterogeneous graph. Utilizing GNNs on heterogeneous graphs offers notable advantages over homogeneous counterparts, particularly in the ability to learn type-specific representations for each node and edge type \citep{wang2019heterogeneous,fu2020magnn}. 

\begin{equation} 
\label{eq:het_gnn} 
\bm{x}^{(k)}_i = \sigma\left(\sum_{c \in \mathcal{C}} \sum_{j \in \mathcal{N}_{c}(i)} \beta_{ij}^c \bm{W}_c \bm{x}^{(k-1)}_j\right), 
\end{equation}
where $\mathcal{C}$ represents the set of edge classes, for instance, the set of vehicle classes. $\bm{W}_c$ is a class-specific learnable weight matrix for the relation class $c \in \mathcal{C}$, and $\beta_{ij}^c $ is the attention weight assigned to node $j$ considering its relationship class $c$ with node $i$. The attention weights for each class are computed as:

\begin{equation} 
\label{eq:het_gnn_1}
\beta_{ij}^c = \frac{\exp(\sigma(\bm{a}_c^T[\bm{W}_c \bm{x}^{(k)}_i | \bm{W}_c \bm{x}^{(k)}_j]))}{\sum_{v \in \mathcal{N}_c(i)} \exp(\sigma(\bm{a}_c^T[\bm{W}_c \bm{x}^{(k)}_i | \bm{W}_c \bm{x}^{(k)}_v]))}, 
\end{equation}
where $\bm{a}_c$ is a relation-specific learnable weight vector, $\mathcal{N}_c(i)$ is the set of neighboring nodes connected by relation class $c$ to node $i$. HGNNs allow for the aggregation of features across multiple classes of relationships, each contributing differently to the node embeddings. This mechanism facilitates learning more intricate and nuanced representations, enhancing the model’s performance across diverse graph-based tasks. In the following sections, we will leverage the expressiveness of the heterogeneous graph neural network to estimate the traffic flow performance of multiple classes of vehicles.

\section{Multi-View Heterogeneous Graph Attention Network for Traffic Assignment}
\label{sec:architecture}
\begin{figure}[!hbt]
  \includegraphics[width=\linewidth]{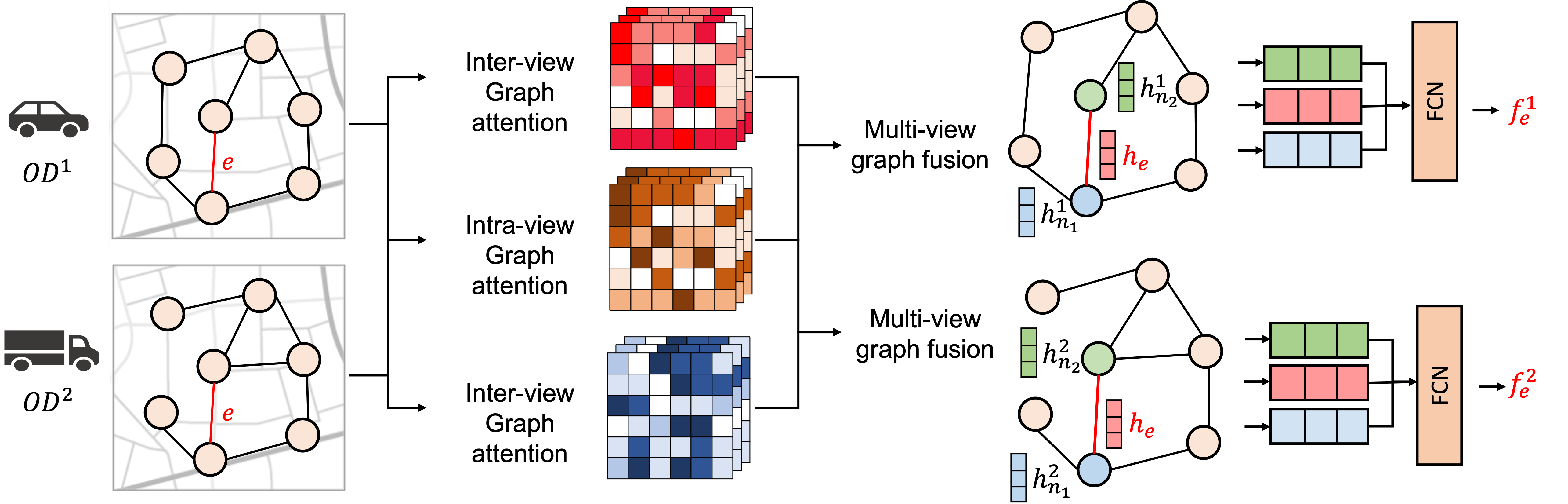}
  \caption{A schematic of the multi-view heterogeneous graph attention network for traffic assignment. The proposed model consists of three parts: (1) graph construction and feature preprocessing module, (2) multi-view graph attention, and (3) multi-view graph fusion. The flow-capacity ratio and link flow of each link are calculated using the source node feature, destination node feature, and normalized edge feature.}
  \label{fig:model}
\end{figure}

In this section, we introduce the proposed architecture of the multi-view heterogeneous graph attention network (referred to heartoafter as M-HetGAT) for traffic assignment. A schematic of the proposed model is shown in Fig. \ref{fig:model}. Each view in the proposed graph neural network represents the traffic flow pattern belonging to one vehicle class. In the following,  detailed explanation of each module including feature preprocessing, multi-view graph attention, and multi-view graph fusion are provided.

\subsection{Graph Construction \& Feature Preprocessing}
The heterogeneous graph for multi-class vehicle traffic assignment modeling, denoted as $\mathcal{G} = \{\mathcal{G}_c(\mathcal{V}^c, \mathcal{E}_\text{r}^c, \mathcal{E}_\text{v}^c) \mid c = 1, \dots, C \}$,   utilizes a multi-view representation of the transportation network. This multi-view GNN representation is motivated and extended from the single-view GNN model from \cite{liu2024end}. Additionally, compared with \cite{liu2024end}, we consider additional interaction between different classes of vehicle. Each view $\mathcal{G}_c$  in the proposed graph neural network represents the traffic flow pattern of that particular vehicle class $c$. The view corresponding to vehicle class $c$ consists of three distinct components: (1) nodes $\mathcal{V}^c$, which correspond to intersections of roadways; (2) real edges $\mathcal{E}^c_\mathrm{r}$, which represent physical connections between intersections, such as streets or highways; and (3) virtual edges $\mathcal{E}^c_\mathrm{v}$, which are auxiliary links between the origin and destination nodes. The virtual edges between the origin and destination node pairs could compress the number of steps of message-passing for long-range node pairs and discover dependencies between link flows and OD demand. It should be noted that for different views of the graph, the set of real or virtual links may differ. This is because certain classes of vehicles may not have access to certain roadways. Furthermore, the incorporation of virtual links is applied to facilitate enhanced feature updating as an edge augmentation technique. 

In the proposed model, the node feature attribute is denoted as $\bm{X}^c \in \mathbb{R}^{|\mathcal{V}^c| \times (|\mathcal{V}^c|+2)}$, where $\mathcal{V}^c$ represents the node set of the $c^{\mathrm{th}}$ view of the graph. Specifically, each row $\bm{x}_u^c \in \mathbb{R}^{1 \times (|\mathcal{V}^c|+2)}$ within this matrix corresponds to the feature vector of a single node $u \in \mathcal{V}^c$, including the origin-destination demand for vehicle class $c$ as well as geographical coordinates. $\bm{Y}^c_{\mathrm{r}} \in \mathbb{R}^{|\mathcal{E}^c_\mathrm{r}| \times 2}$ represents for edge feature of edge $e \in \mathcal{E}_\mathrm{r}^c $ belonging to vehicle class $c$. Each row in this matrix includes the free-flow travel time and the link capacity for a single link.

However, in many cases, the raw node attributes exhibit sparsity and lack normalization. To resolve these issues, a preprocessing step is implemented to transform the raw attributes into a more compact initial node embedding $\bm{x}^{c, 0}_{u}$ in lower-dimensional space through a fully connected network, which represents the updated node embedding of node $u \in \mathcal{V}$ for vehicle class $c \in \mathcal{C}$. This encoding not only preserves the fundamental characteristics of the data but also maintains its semantic integrity. Additionally, we ensure that the edge features are normalized to stabilize their contribution during the subsequent message-passing phases.

\subsection{Multi-View Heterogeneous Graph Attention Network}
\label{sec:extracting}

As discussed in Section \ref{sec:gnn}, a significant challenge within heterogeneous graphs is the effective propagation of node and edge embeddings, specifically between features from different views. To address this challenge, our model involves a novel multi-view graph attention network that facilitates feature extraction and propagation. This network distinguishes between two types of dependencies critical to multi-class vehicle traffic assignment: intra-class and inter-class dependencies. We utilize a graph transformer-based attention mechanism to enhance feature modeling for intra-class dependencies, focusing on the interactions within the same group of node embeddings. On the other hand, for inter-class dependencies, we explore the interactions among different groups of node embeddings, which is especially critical given the complex interactions modeled in multi-class systems.

To address intra-class dependencies, we build an intra-view graph attention layer for each vehicle class. This involves transforming node features into matrices representing keys, queries, and values to compute attention scores among node pairs. The mathematical formulation for vehicle class  $c$ can be expressed as:

\begin{equation}
\begin{aligned}
\label{eq:attention_part1}
\bm{q}_{u,i}^{c,L}, \bm{k}_{u,i}^{c,L}, \bm{v}_{u,i}^{c,L} & = \bm{x}^{c,L}_{u} \ [\bm{W}_{q,i}^{L}, \bm{W}_{k,i}^{L}, \bm{W}_{v,i}^{L}], & \forall u \in \mathcal{V}^c\\
\end{aligned}
\end{equation}
where $\bm{x}^{c,L}_{u}$ is the feature embedding of node $u \in \mathcal{V}$ at the $L^{\mathrm{th}}$ layer. $\bm{q}_{u,i}^{c,L} \in \mathbb{R}^{d_L}$, $\bm{k}_{u,i}^{c,L} \in \mathbb{R}^{d_L}$, and $\bm{v}_{u,i}^{c,L} \in \mathbb{R}^{d_L}$ is the query, key, and value vector at $i^{\mathrm{th}}$ head in encoder where $d_L$ denotes the dimensionality of the feature vectors. $\bm{W}$ are the learnable parameters. We then compute the attention scores among nodes and update the node embedding. Specifically, we first update the node embedding through the virtual link set, then the real link set. Additionally, we incorporate a learnable adaptive weight in computing these scores. This addition of edge-dependent adaptive weights stems from the motivation that variations in low-dimensional node embeddings reflect the variability in node OD demand. In particular, OD pairs are assigned higher attention scores due to their higher influence on traffic flow distribution. By implementing adaptive weights at the edge level, our graph attention model allows for the node feature propagation with the most relevant information via edges,  which enhances the model’s contextual comprehension. Mathematically it could be described as follows

\begin{equation}
\begin{aligned}
\label{eq:attention_part3}
p_{e,i}^{c,L} &= \mathrm{exp} \left(\frac{\bm{q}_{u,i}^{c,L} \bm{k}_{v,i}^{c,L}}{\sqrt{d_L}} \alpha_{e}^{c} \right), & \forall e=(u,v) \in \mathcal{E}_\mathrm{v}^c  \\
\bm{r}^{c,L}_{u,i} &= \left( \sum_{v \in \mathcal{N}_o(u)} p_{e,i}^{c,L} \bm{v}_{v,i}^{c,L} \right) / \left( \sum_{v \in \mathcal{N}_o(u)} p_{e,i}^{c,L} \right), & \forall e=(u,v) \in \mathcal{E}_\mathrm{v}^c\\
s_{e,i}^{c,L} &= \mathrm{exp} \left(\frac{\bm{r}_{u,i}^{c,L} \bm{k}_{v,i}^{c,L}}{\sqrt{d_L}} \beta_{e}^{c} \right), & \forall e=(u,v) \in \mathcal{E}_\mathrm{r}^c  \\
\bm{z}^{c,L}_{u,i} &= \left( \sum_{v \in \mathcal{N}_o(u)} s_{e,i}^{c,L} \bm{v}_{v,i}^{c,L} \right) / \left( \sum_{v \in \mathcal{N}_o(u)} s_{e,i}^{c,L} \right), & \forall e=(u,v) \in \mathcal{E}_\mathrm{r}^c
\end{aligned}
\end{equation}
where $\alpha_{e}^{c} \in \mathbb{R}$ and $\beta_{e}^{c} \in \mathbb{R}$ represent the learnable adaptive weight. $s_{e,i}^{c} \in \mathbb{R}$ represents the un-normalized attention score of the edge $e=(u, v) \in \mathcal{E}$ at $i^{\mathrm{th}}$ head of the $L^{\mathrm{th}}$ layer; $\bm{z}^{c,L}_{u,i}$ represents the intra-class normalized node embedding of node $u$ and $\mathcal{N}_o(u)$ represents all the outgoing nodes connected to $u$.

Furthermore, to model interdependencies between different classes, we propose an inter-view graph attention layer. The structure of the inter-view graph attention layer employed in the model is analogous to Equation \ref{eq:attention_part1}; However, it utilizes a distinct node embedding as the input. Specifically, we concatenate node embeddings from various vehicle classes to generate new inter-class key, query, and value vectors:

\begin{equation}
\begin{aligned}
\label{eq:attention_part2}
\tilde{\bm{q}}_{u,i}^{c,L}, \tilde{\bm{k}}_{u,i}^{c,L}, \tilde{\bm{v}}_{u,i}^{c,L} & = \left[\bm{x}^{1,L}_{u} | \dots | \bm{x}^{C,L}_{u}\right] \ [\tilde{\bm{W}}_{q,i}^{L} \tilde{\bm{W}}_{k,i}^{L}, \tilde{\bm{W}}_{v,i}^{L}], & \forall u \in \mathcal{V}^c\\
\end{aligned}
\end{equation}

These two distinguished node embeddings are crucial for accurately evaluating and forecasting the interactions among different vehicle classes within a network. Similarly, we can calculate the inter-class node embedding $\tilde{\bm{z}}^{c,L}_{u,i}$ using Equation \ref{eq:attention_part3} by using the inter-class key, query, and value matrix. Additionally, a residual connection supplements the final output of the layer, ensuring the integration of original input features with learned representations for enhanced model performance. Subsequently, value vectors, normalized by the attention scores, pass through a position-wise feed-forward network (denoted by $\mathtt{FFN}$), which includes layer normalization (denoted by $\mathtt{LayerNorm}$). Finally, the output of the multi-view graph attention network is obtained by concatenating the inter-view and intra-view node embeddings from all attention heads, according to 

\begin{equation}
\label{eq:virtual_attention}
\begin{aligned}
\bm{x}_{u,i}^{c,L+1} & = \bm{x}_{u,i}^{c,L+1} + \mathtt{LayerNorm} \left(\mathtt{FFN} \left(\bm{z}^{c,L}_{u,i}; \bm{W}_{z}, \bm{b}_{z}\right) + \mathtt{FFN} \left(\tilde{\bm{z}}^{c,L}_{u,i}; \bm{W}_{\tilde{z}}, \bm{b}_{\tilde{z}}\right)\right), & \forall u \in \mathcal{V}^c\\
\bm{x}_u^{c,L+1} & = \left[\bm{x}_{u,0}^{c,L+1} | \bm{x}_{u,1}^{c,L+1} | \dots | \bm{x}_{u,N_h}^{c,L+1} \right], & \forall c \in \mathcal{C}
\end{aligned}
\end{equation}
where $N_h$ represents the number of attention heads of the graph attention layer. To enhance the propagation of node features throughout the network, we sequentially stack multiple layers of the graph attention layer. The final output of this stacked graph attention network, denoted as $\bm{H}^c$, serves as the input for the subsequent edge-level link flow prediction.

\subsection{Graph Edge Prediction}
To predict the traffic flow at the edge level, the node embedding of the source node and destination node, and the normalized real edge feature are concatenated and passed through a feed-forward neural network. In this paper, we predict the flow-capacity ratio, denoted by $\tilde{\alpha}_e$, which is the link flow normalized by the link capacity, according to

\begin{equation}
\label{eq:edge_prediction}
\tilde{\alpha}_e^c = \texttt{FFN}([\bm{h}_{u}^c | \bm{h}_{v}^c | \bm{y}_{e}^{c}]; \bm{W}_h, \bm{b}_h),\quad \forall e=(u,v) \in \mathcal{E}^c_\mathrm{r}
\end{equation}
where $\bm{h}_{u}^c$ and $\bm{h}_{v}^c$ represent the node embeddings, $\bm{x}_{e_\text{r}}^{c}$ is the edge features for the real edge $e$ belonging to vehicle class $c$, $\bm{W}_h$ and $\bm{b}_h$ are the learnable parameters of the feed forward network. The predicted link flows, $\tilde{f}_e^c$, for vehicle class $c$ can be calculated by multiplying the link capacity with the predicted flow-capacity ratio. Subsequently, selecting an appropriate loss function becomes crucial to ensure the model's effective convergence. The proposed model employs a composite loss function comprising two components. The first part is the supervised loss which measures the difference between prediction and ground truth values. It consists of two parts: (1) the discrepancy in the flow-capacity ratios, captured by $L_{\alpha}$,  and (2) the difference in the link flows, captured by $L_{f}$. Thus, we will have
\begin{equation}
\begin{split}
 L_{\alpha} &= \sum_{c \in \mathcal{C}} \sum_{e \in \mathcal{E}^c_\mathrm{r}} \frac{1}{|\mathcal{E}^c_\mathrm{r}|} \|{\alpha}^c_e - \tilde{\alpha}^c_e \| \\
 L_{f} &=   \sum_{c \in \mathcal{C}} \sum_{e \in \mathcal{E}^c_\mathrm{r}} \frac{1}{|\mathcal{E}^c_\mathrm{r}|} \|f^c_e - \tilde{f}^c_e \| , 
    \label{eq:ls}
\end{split}
\end{equation}
where the ${\alpha}^c_e$ and $\tilde{\alpha}^c_e$ represent the ground truth and prediction of flow-capacity ratio on link $e \in \mathcal{E}$ for vehicle class $c$. The $f^c_e$ and $\tilde{f^c_e}$ represent the ground truth and prediction of link flow on link $e \in \mathcal{E}$ for vehicle class $c$. The second part of the loss function originates from the principle of node-based flow conservation, where the total flow of traffic entering a node equals the total flow of traffic exiting that node. The node-based flow conservation law can be represented mathematically by

\begin{equation}
    \sum_k f^c_{ki} - \sum_j f^c_{ij} = \Delta f^c_i = \begin{cases}
 \quad \sum_{v\in\mathcal{V}^c_{OD}} O^c_{v,i} - \sum_{v\in\mathcal{V}^c_{OD}} O^c_{i,v}, & \text{ if } i \in \mathcal{V}^c_{OD}, \\ 
 \quad 0 & \text{ otherwise }, 
\end{cases}
\end{equation}
where $f^c_{ki}$ denotes the flow on the link $(k, i)$, $\Delta f^c_i$ represents the difference between flow receiving and sending at node $i$, $O^c_{v, i}$ represents the number of OD demand from $v$ to $i$. $\mathcal{V}^c_{OD}$ denote the origin-destination node set. The node-based flow conservation law can be added as a  regularization term in the total loss function, thereby promoting compliance with  flow conservation. In order to incorporate the conservation law into the loss function, we form a conservation loss function given by
\begin{equation}
    L_c = \sum_{i} \ |\sum_{k \in \mathcal{N}_{i}(i)} \tilde{f}^c_{ki} - \sum_{j \in \mathcal{N}_{o}(i)} \tilde{f}^c_{ij} - \Delta f^c_i |,
    \label{eq:lf}
\end{equation}
where $\mathcal{N}_{i}(i)$ and $\mathcal{N}_{o}(i)$ represent the incoming and outgoing edges for node $i$, respectively. Consequently. the total loss for the flow prediction $L_{total}$ is the weighted summation of the supervised loss and the conservation loss:

\begin{equation}
    L_{\text{total}} = w_{\alpha} L_{\alpha} + w_{f} L_{f} + w_c L_c,
    \label{eq:total_loss}
\end{equation}
where the $w_{\alpha}$, $w_f$ and $w_c$ represent the normalized weights for the flow-capacity ratio supervised loss, link flow supervised loss, and the conservation loss, respectively. 

\section{Numerical Experiments}
\label{sec:experiment}

In this work, we consider two numerical experiments to evaluate the accuracy, efficiency, and generalization capability of our proposed GNN model. The first experiment focuses on urban transportation networks under normal conditions, while the second experiment models these networks under scenarios involving partial road closures.

\subsection{Experiments on Urban Transportation Networks}
\label{sec:urban}

\subsubsection{Characteristics of networks}
This paper examines three urban transportation networks as case studies: the Sioux Falls network, East Massachusetts Network (EMA), and Anaheim network (see Table \ref{tab:urban_network_detail} for their network information). Detailed information regarding the topology of these networks, characteristics of their links, and the OD demand was obtained from \citep{bar2021transportation}.  In this problem, two vehicle classes are considered: cars and trucks.

\begin{table}[htb!]
\renewcommand{\arraystretch}{1.15}
\centering
\caption{Network information for the three urban transportation networks used in this study}
\label{tab:urban_network_detail}
\begin{tabular}{cccccc}
\hline\hline
Network Name & $|\mathcal{V}|$ & $|\mathcal{E}_\text{r}|$ & Average degree & Aggregated OD Demand \\ \hline
Sioux Falls  & 24      & 76      & 3.17           & 316,290          \\
EMA          & 74      & 258     & 3.49           & 226,168          \\
Anaheim      & 416     & 914    & 3.05           & 311,270          \\ \hline\hline
\end{tabular}%
\end{table}

Furthermore, in order to reflect the larger space occupancy of trucks and their impact on  traffic flow, we adopt a passenger car equivalent ratio equal to 1.9, which considers both the heavy-duty and light-duty trucks \cite{pulugurtha2022passenger}. To create variations from the perspective of both demand and the link capacity, we scaled the demand by a scaling factor according to
\begin{equation}
\label{eq:od_demand}
    \tilde{O}_{s,t} = \delta^o_{s,t} \ O_{s,t},
\end{equation}
where $O_{s,t}$ is the default OD demand between source $s$ and destination $t$ and $\delta^o_{s,t} \sim U(0.5, 1.5)$ is the uniformly distributed random scaling factor for the OD pair ($s$, $t$). Additionally, to account for variations in network properties, variable link capacities are created according to 
\begin{equation}
\label{eq:ca_demand}
    \tilde{c}_{a} = \delta^c_{a} \ c_{a} ,
\end{equation}
where $c_a$ is the original link capacity for link $a$, and $\delta^c_{a}$ is the scaling factor for link $a$. Capacity variations are considered to be due to traffic accidents, road construction/damage, and adverse weather conditions, which reduce the link capacity. In this work, the level of capacity reduction is randomly drawn from a uniform distribution between 0.8 and 1, i.e., $\delta^c_{a} \sim U(0.8, 1.0)$. 

\subsubsection{Experiment setup}
\label{sec:setup}
The dataset of each network includes 5,000 entries. These entries are divided into training and testing subsets by 80\% and 20\% of the dataset. To ensure the robustness of the model, it is critical to confirm that the distribution of datasets reflects realistic variations in network conditions and demand patterns. The lowest observed coefficient of variation for link capacity and OD demand across the networks is 0.45 and 0.22, respectively. This variation suggests that the datasets are adequately diverse, enabling the exploration of different scenarios.

The training and testing datasets were generated using the Frank-Wolfe algorithm to solve User Equilibrium Traffic Assignment Problems (UE-TAP) and System Optimal Traffic Assignment Problems (SO-TAP) \citep{fukushima1984modified}. Convergence of the algorithm is achieved when the square root of the sum of squared differences in link flows between two consecutive iterations, normalized by the total link flow, is below a threshold of 1e-5. The GNN model is developed in PyTorch \citep{paszke2019pytorch} and the Deep Graph Library (DGL) \citep{wang2019deep}, featuring a preprocessing layer that includes a three-layer fully connected neural network with an embedding dimension of 32. The model architecture consists of four multi-view graph attention layers. The number of heads in the attention mechanism is 8. Hyperparameters were carefully selected, setting the hidden layer size at 64, which is common in neural network configuration \citep{liu2023physics1}. Training parameters were set with a learning rate of 0.001 and a batch size of 128. The loss function weights, corresponding to $L_\alpha$, $L_f$, and $L_c$ in equation \ref{eq:total_loss}, were set at 1.0, 0.005, and 0.05, respectively, to balance the model's training emphasis on different error components.

We conducted a performance evaluation of our proposed multi-view heterogeneous graph attention model (referred to by M-HetGAT), and compared it against three alternative models incorporating multi-view architectures: the multi-view graph attention network (M-GAT), the multi-view graph convolution network (M-GCN), and the multi-view inductive graph representation learning model (M-GraphSAGE). The M-GAT, M-GCN, and M-graphSAGE implementations consist of four layers of graph message passing followed by three fully connected layers, each with an embedding size of 64. We assessed model performance using two key metrics, including mean absolute error (MAE) and root mean square error (RMSE):

\begin{equation}
\mathrm{MAE} = \frac{1}{N}\sum_{i=1}^{N}|y_i-\tilde{y}_i|,
\end{equation}

\begin{equation}
\mathrm{RMSE} = \sqrt{\frac{1}{N}\sum_{i=1}^{N}(y_i-\tilde{y}_i)^2},
\end{equation}

where $y$ and $\tilde{y}$ respectively represent the ground truth and predicted values for quantity of interest. We conducted a 5-fold cross-validation for each experiment to ensure the robustness of our results across different subsets of the data.

\subsubsection{Numerical results}

\begin{figure}[htb!]
\centering
\begin{subfigure}[b]{0.24\textwidth}
    \centering
    \includegraphics[width=\textwidth]{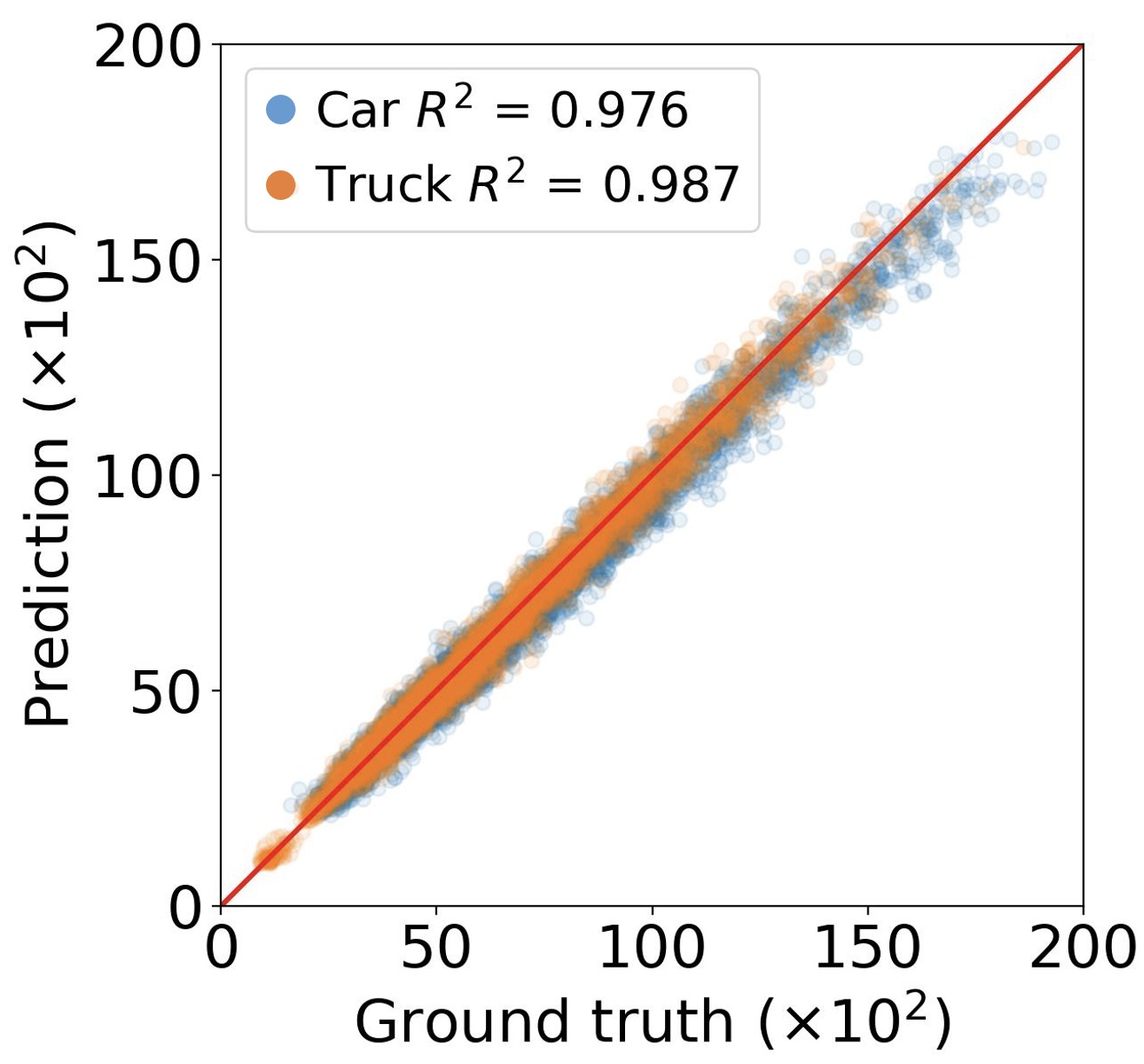}
    \caption{Ours}
    \label{fig:correlation_1}
\end{subfigure}
\hfill
\begin{subfigure}[b]{0.24\textwidth}
    \centering
    \includegraphics[width=\textwidth]{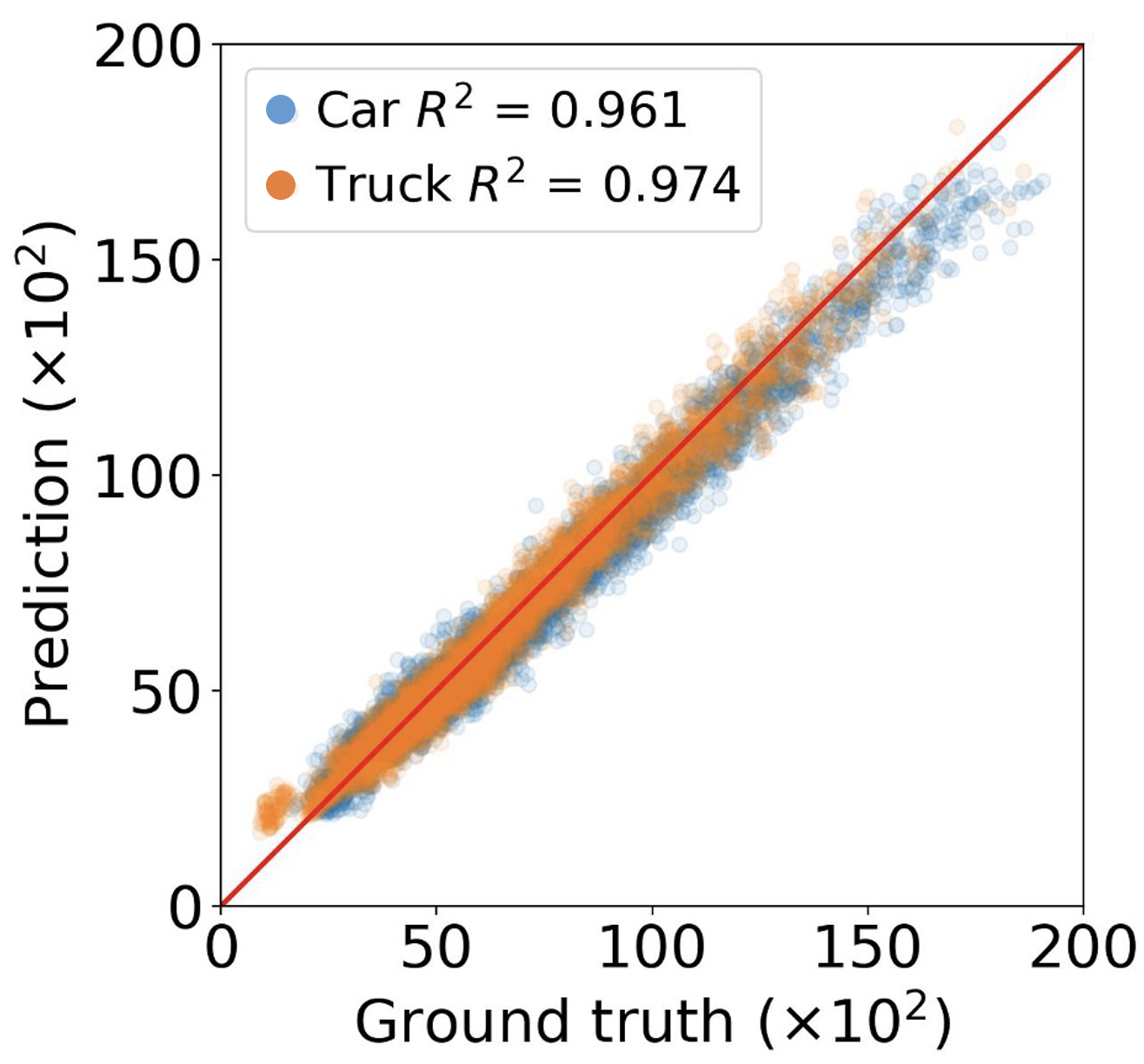}
    \caption{GAT}
    \label{fig:correlation_2}
\end{subfigure}
\hfill
\begin{subfigure}[b]{0.24\textwidth}
    \centering
    \includegraphics[width=\textwidth]{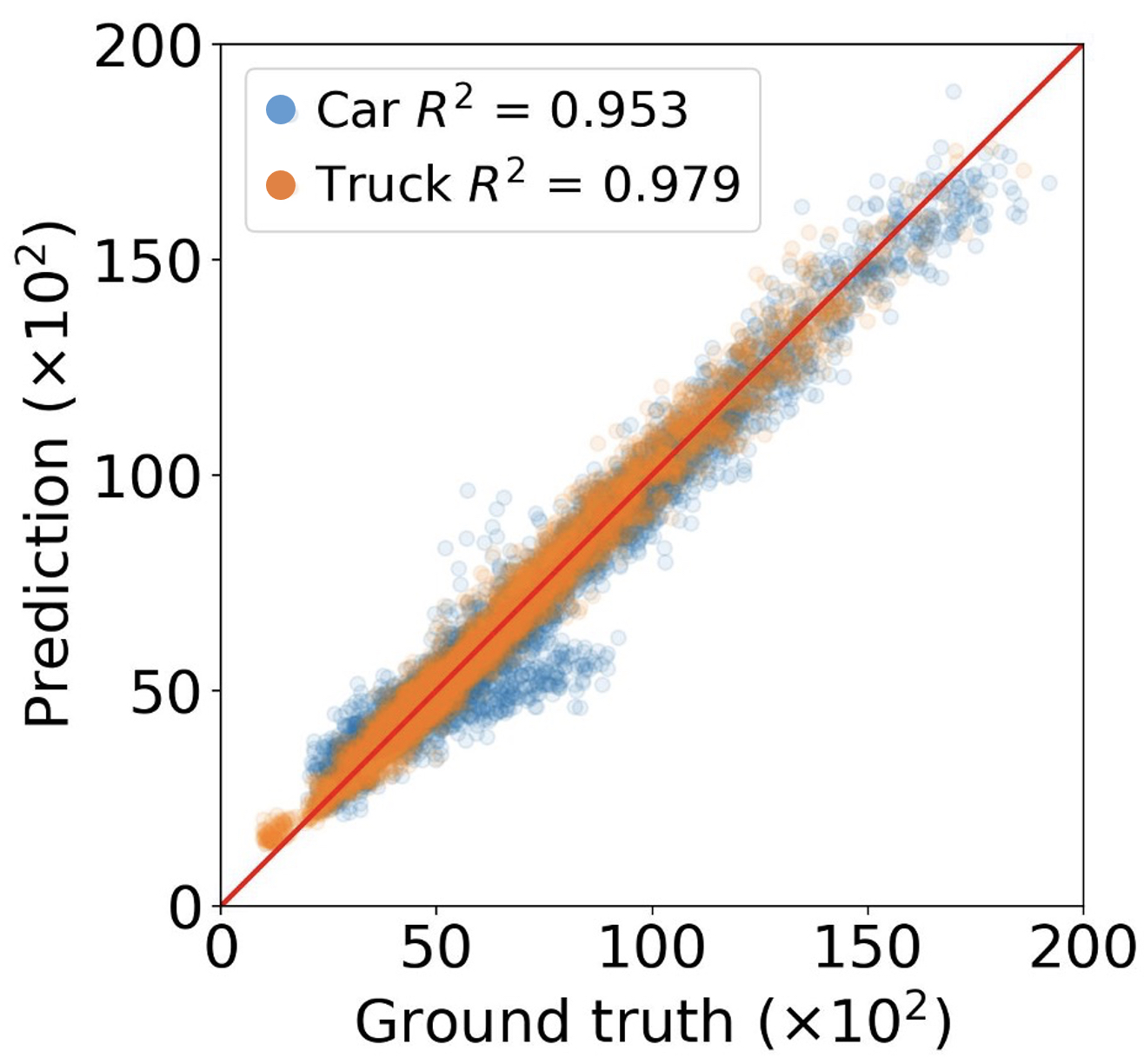}
    \caption{GCN}
    \label{fig:correlation_3}
\end{subfigure}
\hfill
\begin{subfigure}[b]{0.24\textwidth}
    \centering
    \includegraphics[width=\textwidth]{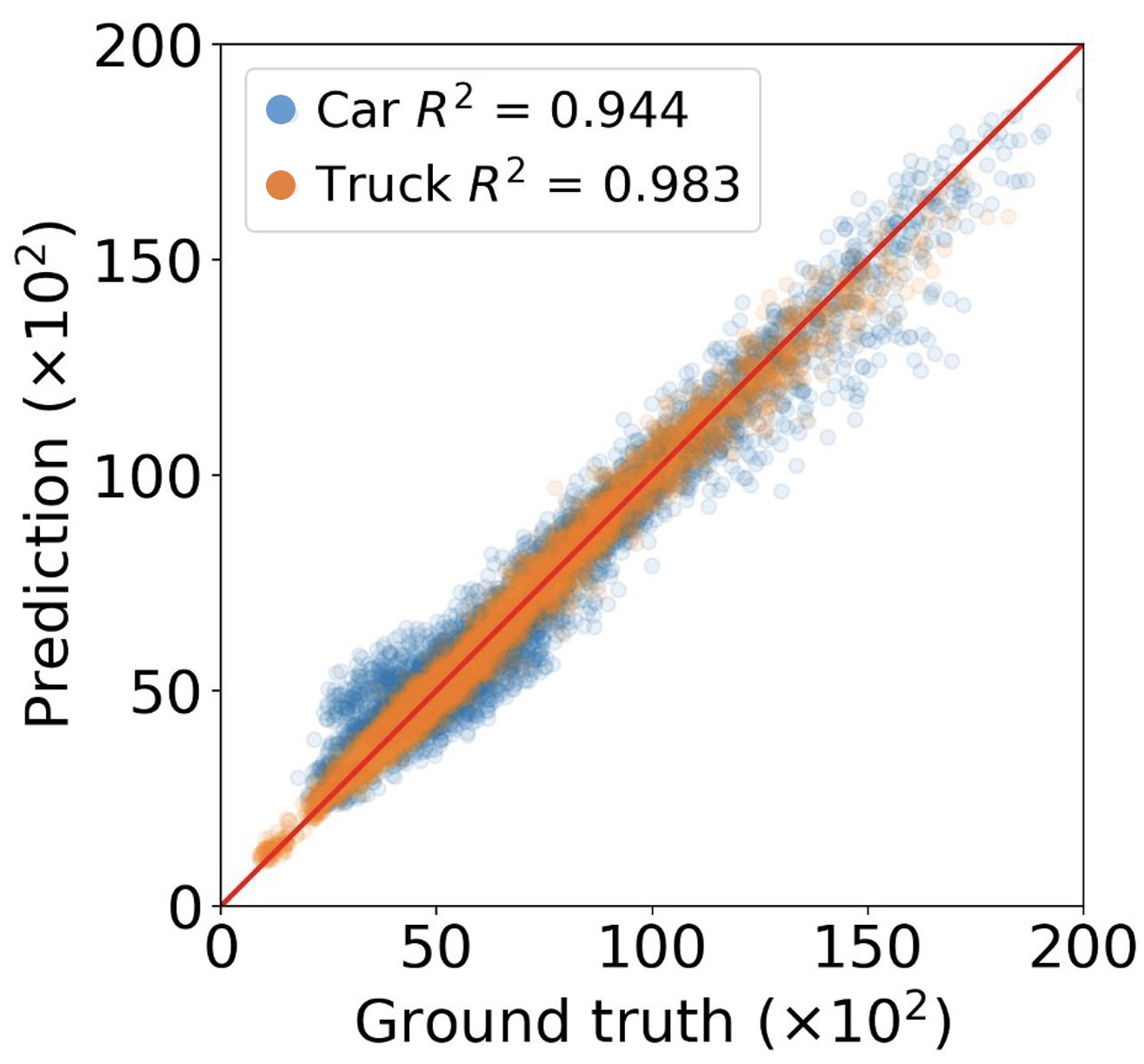}
    \caption{GraphSAGE}
    \label{fig:correlation_4}
\end{subfigure}
    \caption{Comparison of predicted versus ground truth link flows in the Sioux Falls network, using four different models.}
    \label{fig:correlation}
\end{figure}

The experiments under the urban road network are conducted in two different settings: the system optimal (SO-TAP) and user equilibrium (UE-TAP) traffic assignment problems. Figure~\ref{fig:correlation} illustrates the correlation between the predicted and actual link flows in the Sioux Falls network for both cars and trucks. The analysis involves predicting the flow-capacity ratios for 10,000 edges using various models: M-HetGAT, M-GAT, M-GCN, and M-graphSAGE. The correlation coefficients, exceeding 0.94 as depicted in Figure~\ref{fig:correlation}, underscore the effectiveness of the multi-view graph attention mechanisms in accurately forecasting link flows and utilization across different vehicle classes. To illustrate, The inter-view message passing helps the GNN model to understand the interaction between different vehicle classes. On the other hand, the intra-view message passing would help the single-view GNN model to predict the traffic flow for each class of vehicles accurately. Moreover, the results reveal that the heterogeneous graph attention model, M-HetGAT, consistently surpasses the performance of other models such as M-GAT, M-GCN, and M-graphSAGE in pairwise comparisons, demonstrating its superior capability in handling multi-class vehicle flow predictions on urban road networks.

\begin{table}[hbt!]
\centering
\caption{Performance comparison of the proposed model with benchmark methods, on Sioux Falls, East Massachusetts, and Anaheim networks. The mean absolute error and root mean square error are used to evaluate the prediction performance on the testing set.}
\label{tab:urban_orginal}
\renewcommand{\arraystretch}{1.15}
\resizebox{\textwidth}{!}{%
\begin{tabular}{ccccccc|cccc|cccc}
\hline
\multirow{3}{*}{Setting} & \multirow{3}{*}{Metric} & \multirow{3}{*}{Model} & \multicolumn{4}{c}{Sioux   Falls} & \multicolumn{4}{c}{EMA} & \multicolumn{4}{c}{Anaheim} \\ \cline{4-15} 
 &  &  & \multicolumn{2}{c}{Car} & \multicolumn{2}{c}{Truck} & \multicolumn{2}{c}{Car} & \multicolumn{2}{c}{Truck} & \multicolumn{2}{c}{Car} & \multicolumn{2}{c}{Truck} \\ \cline{4-15} 
 &  &  & MAE & RMSE & MAE & RMSE & MAE & RMSE & MAE & RMSE & MAE & RMSE & MAE & RMSE \\ \hline
\multirow{8}{*}{\begin{tabular}[c]{@{}c@{}}System\\ optimal\end{tabular}} & \multirow{4}{*}{\begin{tabular}[c]{@{}c@{}}Flow\\ ($1 \times 10^2$)\end{tabular}} & M-GAT & 4.71 & 5.99 & 3.86 & 5.05 & 0.70 & 1.10 & 0.68 & 1.08 & 3.27 & 4.82 & 0.82 & 1.22 \\
 &  & M-GCN & 6.52 & 10.24 & 3.46 & 4.98 & 0.62 & 1.05 & 0.64 & 1.11 & 4.91 & 6.95 & 0.69 & 0.99 \\
 &  & M-GraphSAGE & 4.33 & 5.59 & 2.78 & 3.71 & 0.48 & 0.77 & 0.41 & 0.64 & 3.22 & 4.76 & 0.75 & 1.09 \\
 &  & M-HetGAT & \textbf{2.85} & \textbf{3.71} & \textbf{2.11} & \textbf{2.81} & \textbf{0.28} & \textbf{0.45} & \textbf{0.25} & \textbf{0.40} & \textbf{3.17} & \textbf{4.24} & \textbf{0.44} & \textbf{0.64} \\ \cline{2-15} 
 & \multirow{4}{*}{\begin{tabular}[c]{@{}c@{}}Link\\ utilization\end{tabular}} & M-GAT & 3.57\% & 4.78\% & 2.95\% & 4.38\% & 2.22\% & 3.38\% & 2.15\% & 3.29\% & 6.21\% & 8.71\% & 1.76\% & 2.38\% \\
 &  & M-GCN & 4.78\% & 7.32\% & 2.55\% & 4.03\% & 2.04\% & 3.46\% & 2.24\% & 3.70\% & 6.91\% & 10.50\% & 1.56\% & 2.11\% \\
 &  & M-GraphSAGE & 3.30\% & 4.51\% & 1.99\% & 2.69\% & 1.50\% & 2.45\% & 1.32\% & 2.19\% & 4.64\% & 6.45\% & 1.45\% & 1.90\% \\
 &  & M-HetGAT & \textbf{2.31\%} & \textbf{3.07\%} & \textbf{1.60\%} & \textbf{2.11\%} & \textbf{0.99\%} & \textbf{1.65\%} & \textbf{0.78\%} & \textbf{1.32\%} & \textbf{2.88\%} & \textbf{3.85\%} & \textbf{0.88\%} & \textbf{1.20\%} \\ \hline
\multirow{8}{*}{\begin{tabular}[c]{@{}c@{}}User\\ equilibrium\end{tabular}} & \multirow{4}{*}{\begin{tabular}[c]{@{}c@{}}Flow\\ ($1 \times 10^2$)\end{tabular}} & M-GAT & 4.64 & 5.97 & 3.35 & 4.42 & 0.60 & 0.99 & 0.74 & 1.18 & 3.48 & 5.55 & 0.75 & 1.16 \\
 &  & M-GCN & 4.83 & 6.78 & 3.03 & 4.06 & 0.93 & 1.46 & 0.81 & 1.23 & 4.87 & 7.21 & 1.06 & 1.56 \\
 &  & M-GraphSAGE & 5.37 & 7.22 & 2.70 & 3.62 & 0.44 & 0.75 & 0.44 & 0.76 & 2.65 & 4.03 & 0.78 & 1.20 \\
 &  & M-HetGAT & \textbf{3.22} & \textbf{4.23} & \textbf{2.39} & \textbf{3.26} & \textbf{0.29} & \textbf{0.50} & \textbf{0.29} & \textbf{0.50} & \textbf{2.62} & \textbf{3.72} & \textbf{0.53} & \textbf{0.77} \\ \cline{2-15} 
 & \multirow{4}{*}{\begin{tabular}[c]{@{}c@{}}Link\\ utilization\end{tabular}} & M-GAT & 3.47\% & 4.59\% & 2.55\% & 3.71\% & 1.88\% & 3.44\% & 2.31\% & 3.71\% & 7.43\% & 11.25\% & 1.87\% & 2.59\% \\
 &  & M-GCN & 3.84\% & 5.97\% & 2.13\% & 2.82\% & 3.12\% & 4.94\% & 2.97\% & 4.91\% & 9.53\% & 14.16\% & 2.11\% & 2.95\% \\
 &  & M-GraphSAGE & 4.29\% & 6.08\% & 1.91\% & 2.50\% & 1.30\% & 2.18\% & 1.28\% & 2.25\% & 5.02\% & 7.24\% & 1.30\% & 1.92\% \\
 &  & M-HetGAT & \textbf{2.67\%} & \textbf{3.49\%} & \textbf{1.64\%} & \textbf{2.18\%} & \textbf{0.92\%} & \textbf{1.64\%} & \textbf{0.79\%} & \textbf{1.39\%} & \textbf{1.99\%} & \textbf{2.93\%} & \textbf{0.57\%} & \textbf{0.90\%} \\ \hline
\end{tabular}%
}
\end{table}

Table \ref{tab:urban_orginal} provides a comparative analysis of the predictive accuracy of various models in different scenarios, highlighting that M-HetGAT consistently outperforms its counterparts. This superiority is demonstrated as the graph size expands, where M-HetGAT sustains a lower MAE and RMSE than models based on homogeneous networks. Specifically, within the Sioux Falls network, M-HetGAT achieves flow MAEs that are reduced by 29.1\% and 25.6\% compared to the next best model in system optimal and user equilibrium scenarios, respectively. Similarly, in the Anaheim network, the reductions are pronounced at 12.9\% and 22.1\%. In the EMA network, the reductions are even more pronounced at 41.2\% and 34.1\% under the same settings. These results underscore the effectiveness of integrating multi-view modeling and virtual links into proposed model, which enhances their capacity to accurately capture and predict traffic flow dynamics. The incorporation of these features likely allows for a deeper understanding of interconnected traffic patterns, leading to improved model robustness and reliability in varying network conditions.

\subsection{Experiments on Urban Transportation Networks with Altered Topologies}
\label{sec:synthetic}

In this section, unlike the experiments outlined in Section \ref{sec:urban}, where training and testing were conducted on the network with the same topologies, we investigate how well our proposed M-HetGAT model can perform on networks with different topologies. Specifically, we consider real-world situations where network links might be completely closed due to maintenance work or disasters such as bridge failures, resulting in significant changes to the network topology. These variations can lead to differences in network structures, as noted by \citep{rodrigue2020geography}. Training models for each unique network configuration requires considerable time and resources. Driven by these challenges, our objective is to assess the adaptability of the M-HetGAT model across diverse topologies. In this study, we use the transportation networks from Section \ref{sec:urban}, specifically modified Sioux Falls, EMA, and Anaheim networks where one to three links were randomly removed from the original network but overall connectivity was preserved. Furthermore, We explore two training approaches: (1) in-distribution training, where the model is both trained and tested under the same modified network dataset; (2) out-of-distribution training, where the model is trained with minor disturbances but tested against major disruptions, simulating real-world unexpected scenarios.

\begin{figure}[htb!]
\centering
\begin{subfigure}[b]{0.33\textwidth}
    \centering
    \includegraphics[width=\textwidth]{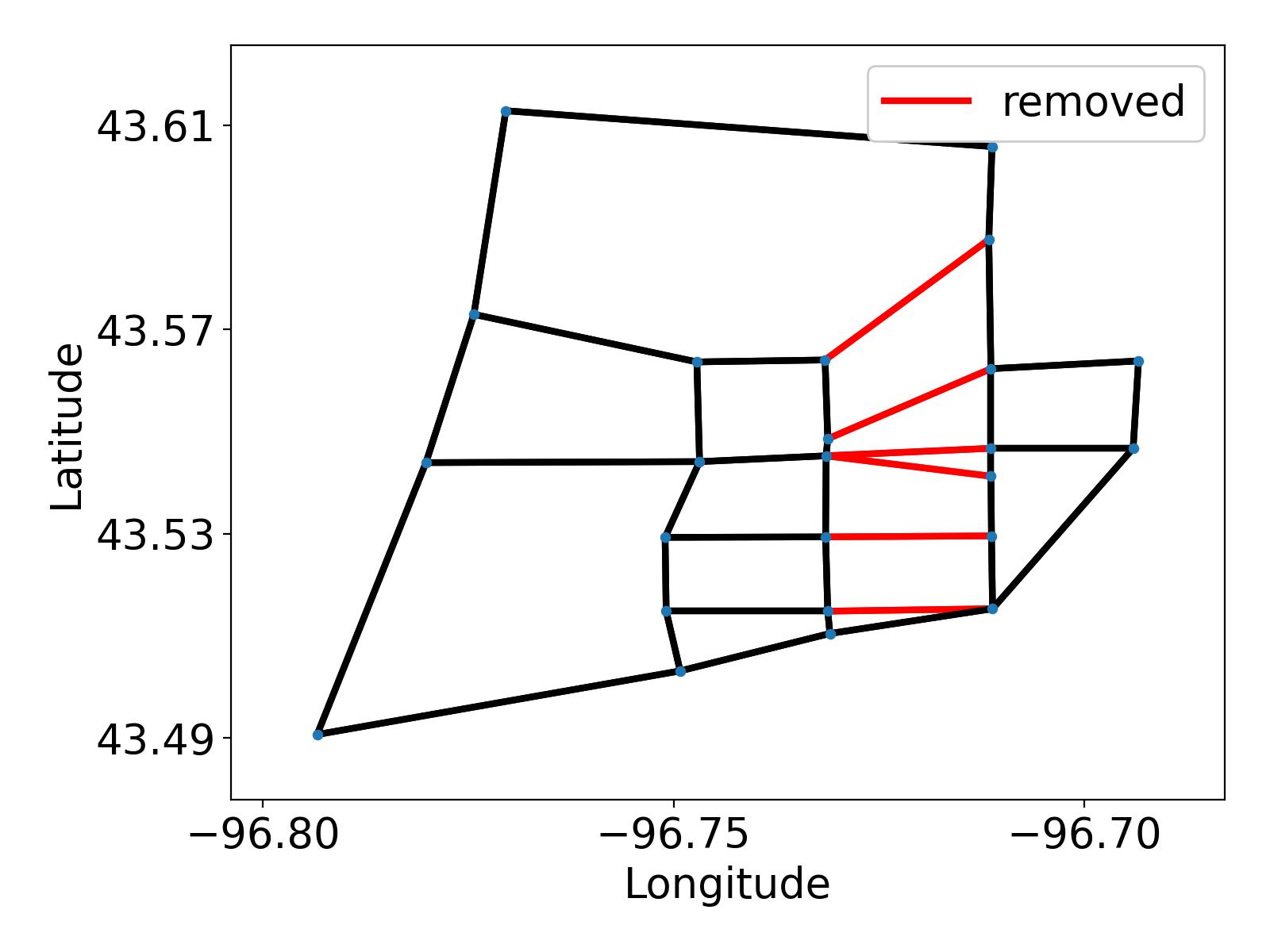}
    \caption{Sioux Falls network}
    \label{fig:Sioux}
\end{subfigure}
\hfill
\begin{subfigure}[b]{0.32\textwidth}
    \centering
    \includegraphics[width=\textwidth]{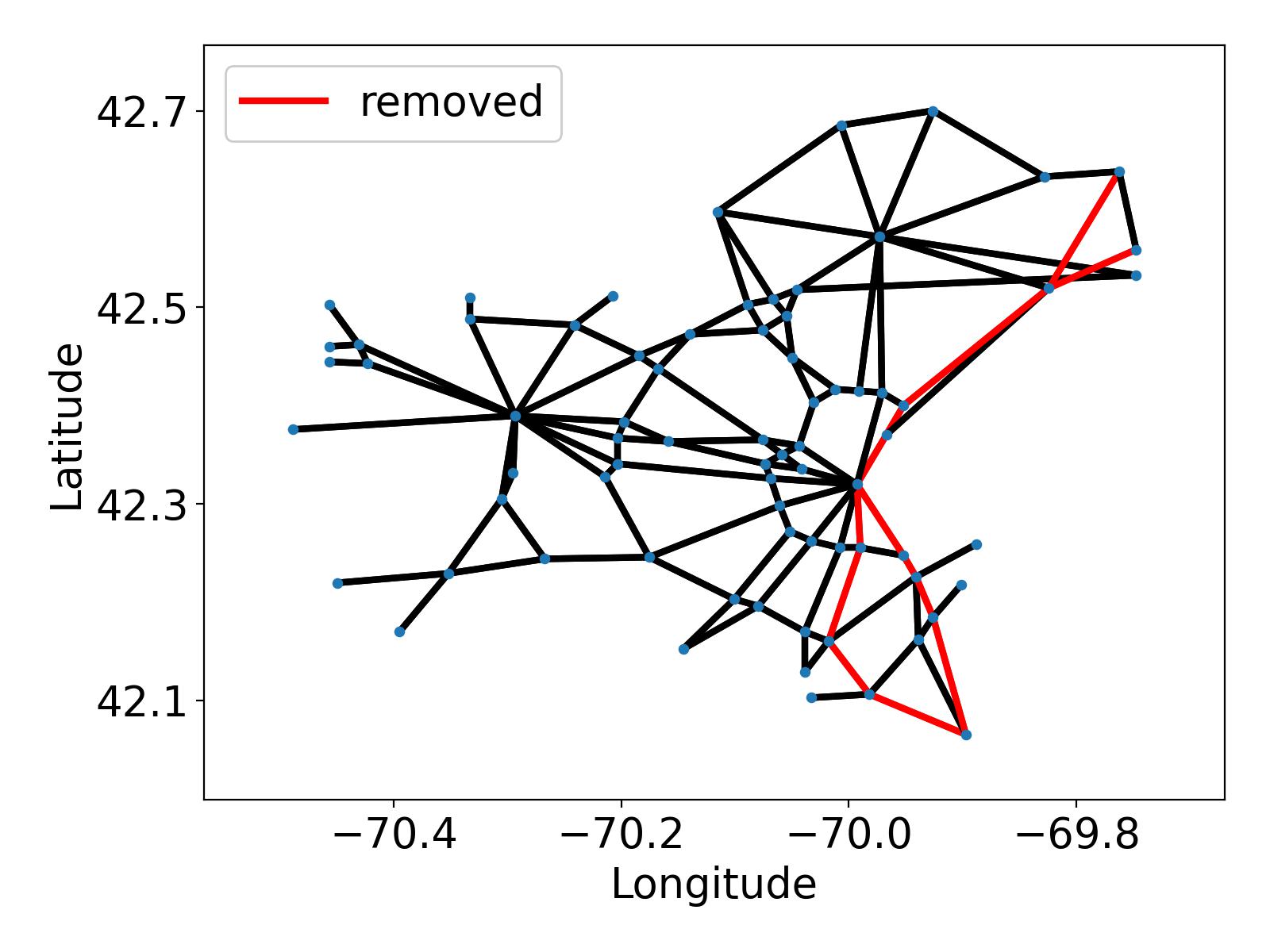}
    \caption{EMA network}
    \label{fig:EMA}
\end{subfigure}
\hfill
\begin{subfigure}[b]{0.33\textwidth}
    \centering
    \includegraphics[width=\textwidth]{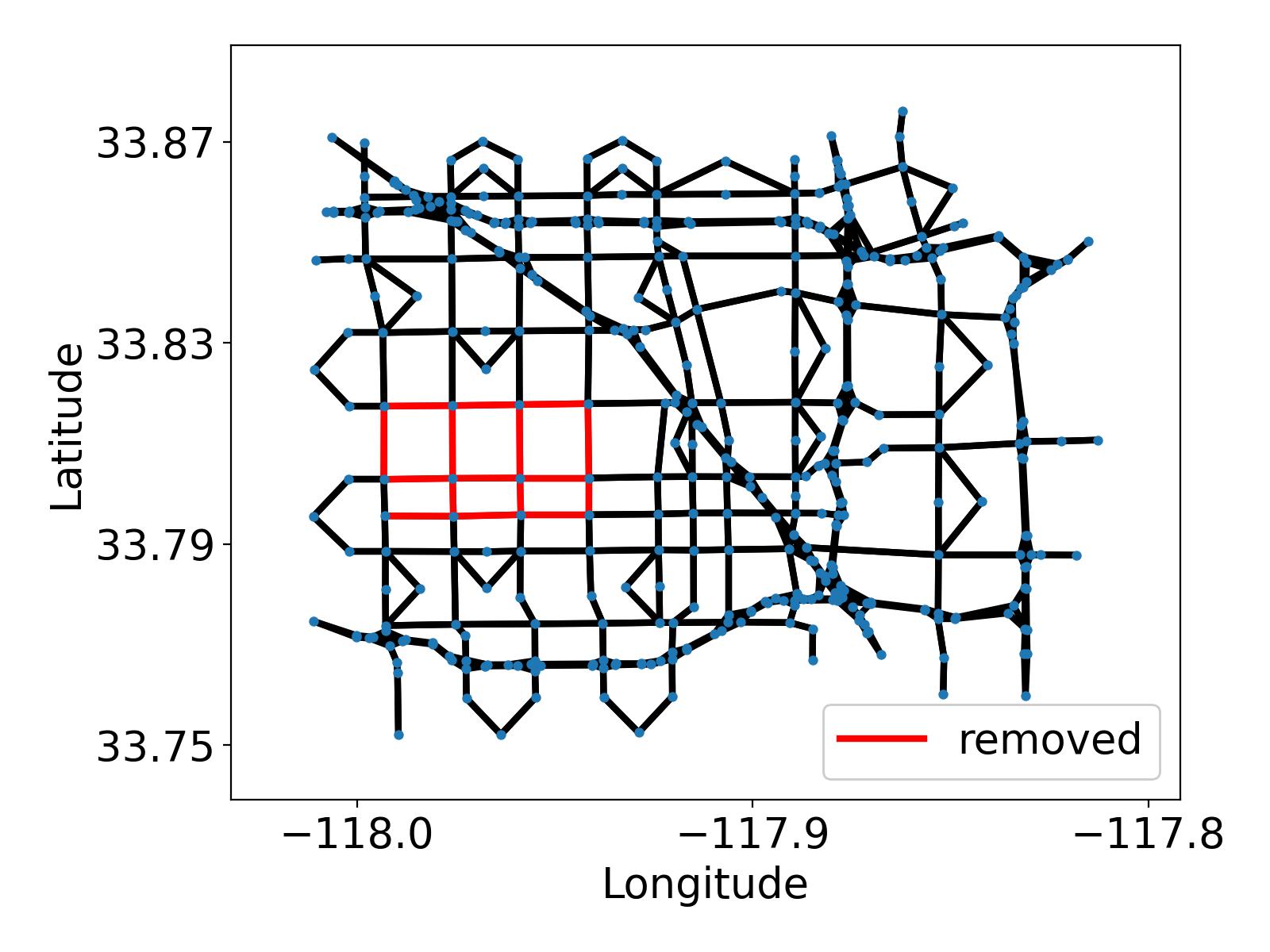}
    \caption{Anaheim network}
    \label{fig:anaheim}
\end{subfigure}
    \caption{The illustration of urban transportation networks with the removed links shown in red.}
    \label{fig:urban_network}
\end{figure}

\subsubsection{In-distribution testing on modified network}
\label{sec:id}
In this experiment, the training and testing datasets both consist of transportation networks where one to three links were randomly removed from a subset of the links. The subset of links is shown in Figure \ref{fig:urban_network}. The link removal will create numerous variations to the original transportation systems, and will reflect the real-world scenarios. The training strategies and model architectures remained unchanged as Section \ref{sec:setup}. As shown in Table \ref{tab:urban_generalization}, a comparative analysis of predictive accuracy across different models and scenarios reveals that M-HetGAT consistently achieves superior performance. For instance, in the Sioux Falls network, M-HetGAT reduces flow MAEs by 23.2\% in the system-optimal scenario and 29.1\% in the user equilibrium scenario, outperforming the next best model. Moreover, the improvements in the EMA network are even more significant, with MAE reductions of 51.2\% and 55.2\%, respectively. These results emphasize the benefits of integrating multi-view modeling and virtual links into the proposed model, which likely enhances its ability to capture and predict complex traffic flow dynamics. Specifically, the model identifies missing links in the networks using a GNN, and the updated node features are processed through both intra-view and inter-view graph attention networks. This approach not only contributes to the model's generalization capabilities but also enhances its robustness and reliability across diverse network conditions.

\begin{table}[htb!]
\centering
\caption{In-distribution testing performance comparison of the proposed model with benchmark methods. Three networks are considered including Sioux Falls, East Massachusetts, and Anaheim. The mean absolute error, root mean square error is used to evaluate the prediction performance on the testing set.}
\label{tab:urban_generalization}
\renewcommand{\arraystretch}{1.15}
\resizebox{\textwidth}{!}{%
\begin{tabular}{ccccccc|cccc|cccc}
\hline
\multirow{3}{*}{Setting} & \multirow{3}{*}{Metric} & \multirow{3}{*}{Model} & \multicolumn{4}{c}{Sioux   Falls} & \multicolumn{4}{c}{EMA} & \multicolumn{4}{c}{Anaheim} \\ \cline{4-15} 
 &  &  & \multicolumn{2}{c}{Car} & \multicolumn{2}{c}{Truck} & \multicolumn{2}{c}{Car} & \multicolumn{2}{c}{Truck} & \multicolumn{2}{c}{Car} & \multicolumn{2}{c}{Truck} \\ \cline{4-15} 
 &  &  & MAE & RMSE & MAE & RMSE & MAE & RMSE & MAE & RMSE & MAE & RMSE & MAE & RMSE \\ \hline
\multirow{8}{*}{\begin{tabular}[c]{@{}c@{}}System\\ optimal\end{tabular}} & \multirow{4}{*}{Flow} & M-GAT & 5.33 & 6.89 & 3.23 & 4.23 & 0.54 & 0.88 & 0.52 & 0.84 & 3.42 & 4.57 & 0.71 & 1.00 \\
 &  & M-GCN & 4.00 & 5.23 & 3.16 & 4.12 & 0.72 & 1.23 & 0.58 & 0.96 & 3.45 & 4.68 & 0.62 & 0.88 \\
 &  & M-GraphSAGE & 4.53 & 5.92 & 3.14 & 4.14 & 0.39 & 0.65 & 0.39 & 0.63 & 2.63 & 3.67 & 0.46 & 0.69 \\
 &  & M-HetGAT & \textbf{3.08} & \textbf{4.22} & \textbf{2.35} & \textbf{3.34} & \textbf{0.24} & \textbf{0.39} & \textbf{0.19} & \textbf{0.33} & \textbf{2.40} & \textbf{3.36} & \textbf{0.33} & \textbf{0.49} \\ \cline{2-15} 
 & \multirow{4}{*}{Ratio} & M-GAT & 4.08\% & 5.75\% & 2.34\% & 3.23\% & 1.72\% & 2.75\% & 1.58\% & 2.51\% & 2.62\% & 3.78\% & \textbf{0.71\%} & \textbf{1.07\%} \\
 &  & M-GCN & 3.03\% & 4.18\% & 2.31\% & 3.08\% & 2.23\% & 3.51\% & 1.98\% & 3.25\% & 2.74\% & 4.14\% & 0.82\% & 1.19\% \\
 &  & M-GraphSAGE & 3.40\% & 4.63\% & 2.29\% & 3.09\% & 1.19\% & 1.93\% & 1.26\% & 2.07\% & 2.78\% & 3.79\% & 0.97\% & 1.25\% \\
 &  & M-HetGAT & \textbf{2.24\%} & \textbf{3.20\%} & \textbf{1.70\%} & \textbf{2.56\%} & \textbf{0.72\%} & \textbf{1.18\%} & \textbf{0.58\%} & \textbf{0.98\%} & \textbf{2.52\%} & \textbf{3.29\%} & 0.88\% & 1.09\% \\ \hline
\multirow{8}{*}{\begin{tabular}[c]{@{}c@{}}User\\ equilibrium\end{tabular}} & \multirow{4}{*}{Flow} & M-GAT & 4.09 & 5.24 & 3.01 & 3.96 & 0.56 & 0.96 & 0.56 & 0.97 & 3.30 & 4.62 & 0.54 & 0.79 \\
 &  & M-GCN & 3.91 & 5.05 & 3.12 & 4.14 & 0.83 & 1.39 & 0.78 & 1.34 & 2.85 & 4.22 & 0.56 & 0.87 \\
 &  & M-GraphSAGE & 4.22 & 5.52 & 3.08 & 4.04 & 0.48 & 0.86 & 0.45 & 0.83 & 3.02 & 4.30 & 0.72 & 1.05 \\
 &  & M-HetGAT & \textbf{3.29} & \textbf{4.25} & \textbf{2.29} & \textbf{2.98} & \textbf{0.26} & \textbf{0.49} & \textbf{0.24} & \textbf{0.46} & \textbf{2.29} & \textbf{3.40} & \textbf{0.44} & \textbf{0.66} \\ \cline{2-15} 
 & \multirow{4}{*}{Ratio} & M-GAT & 3.12\% & 4.21\% & 2.17\% & 2.93\% & 1.70\% & 3.03\% & 1.76\% & 3.27\% & 4.71\% & 6.41\% & 1.48\% & 1.91\% \\
 &  & M-GCN & 2.97\% & 4.00\% & 2.29\% & 3.13\% & 2.47\% & 3.88\% & 2.24\% & 3.37\% & 3.32\% & 4.90\% & 0.93\% & 1.44\% \\
 &  & M-GraphSAGE & 3.29\% & 4.65\% & 2.29\% & 3.14\% & 1.41\% & 2.58\% & 1.35\% & 2.51\% & 2.89\% & 4.17\% & 0.78\% & 1.16\% \\
 &  & M-HetGAT & \textbf{2.48\%} & \textbf{3.40\%} & \textbf{1.68\%} & \textbf{2.29\%} & \textbf{0.73\%} & \textbf{1.35\%} & \textbf{0.68\%} & \textbf{1.34\%} & \textbf{1.64\%} & \textbf{2.54\%} & \textbf{0.44\%} & \textbf{0.64\%} \\ \hline
\end{tabular}%
}
\end{table}

\begin{figure}[htb!]
\centering
\begin{subfigure}[b]{0.3\textwidth}
    \centering
    \includegraphics[width=\textwidth]{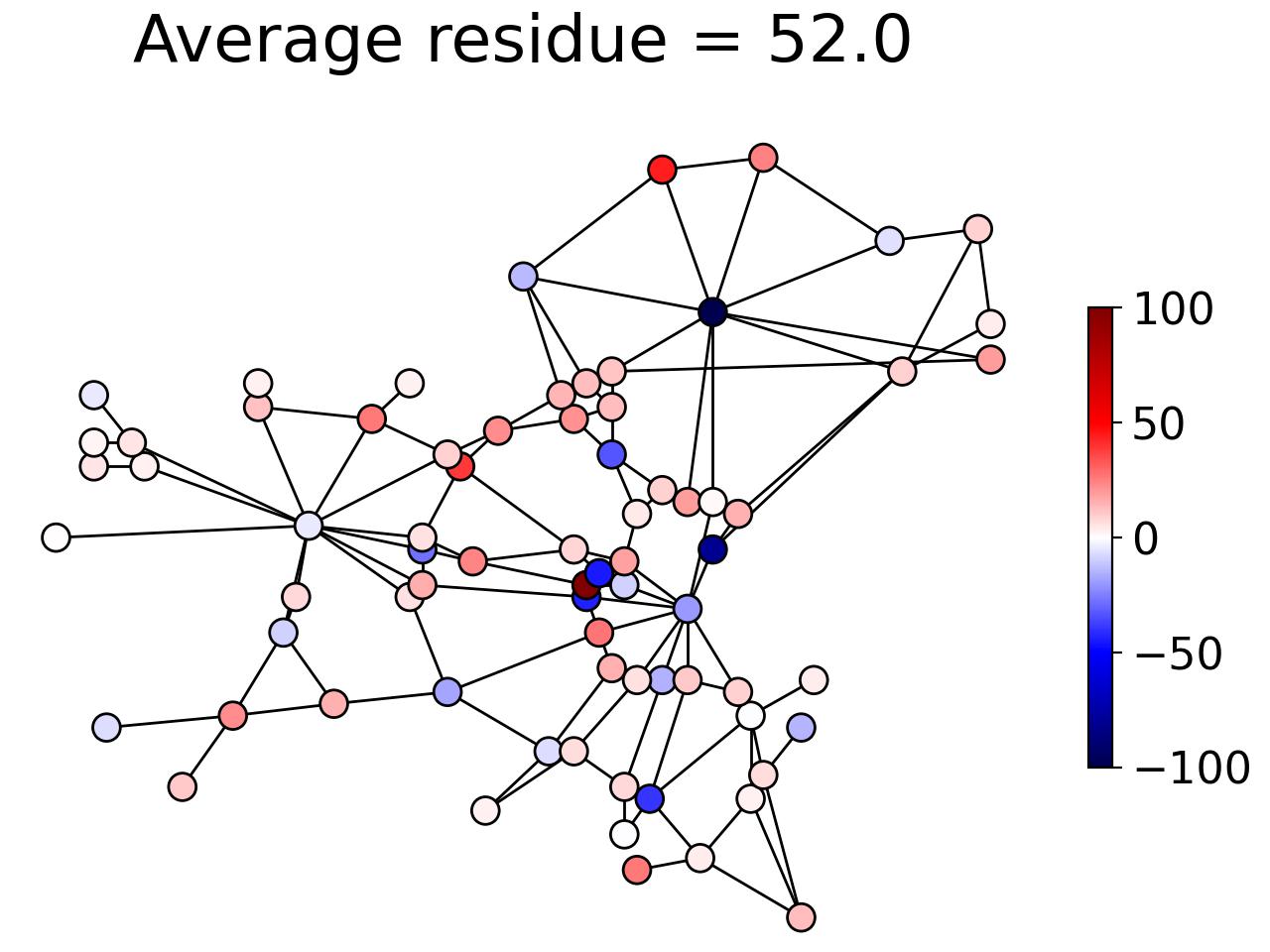}
    \caption{Ours}
    \label{fig:worst_residue_1}
\end{subfigure}
\hspace{5em}
\begin{subfigure}[b]{0.3\textwidth}
    \centering
    \includegraphics[width=\textwidth]{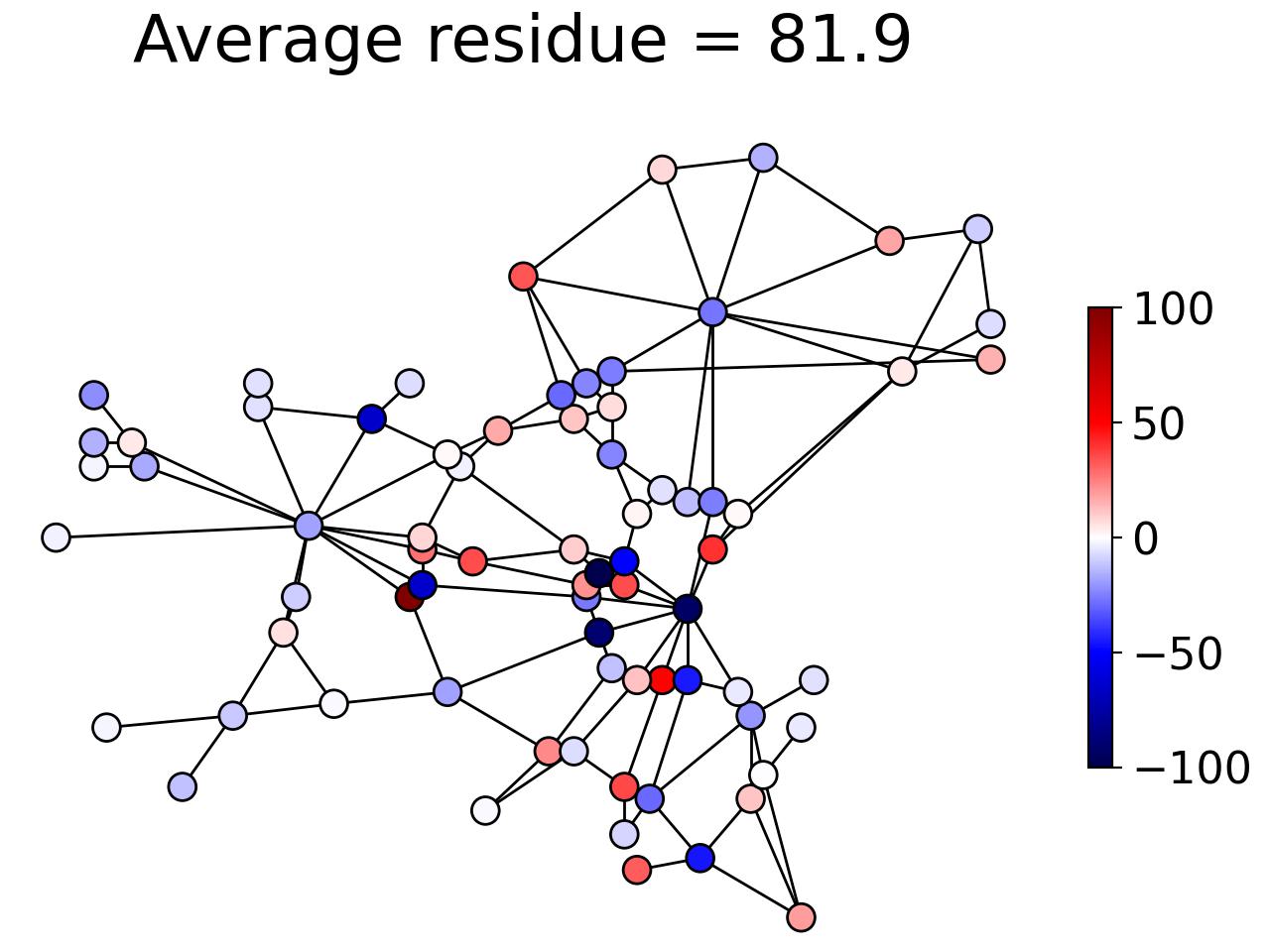}
    \caption{GAT}
    \label{fig:worst_residue_2}
\end{subfigure}

\vspace{1em}

\begin{subfigure}[b]{0.3\textwidth}
    \centering
    \includegraphics[width=\textwidth]{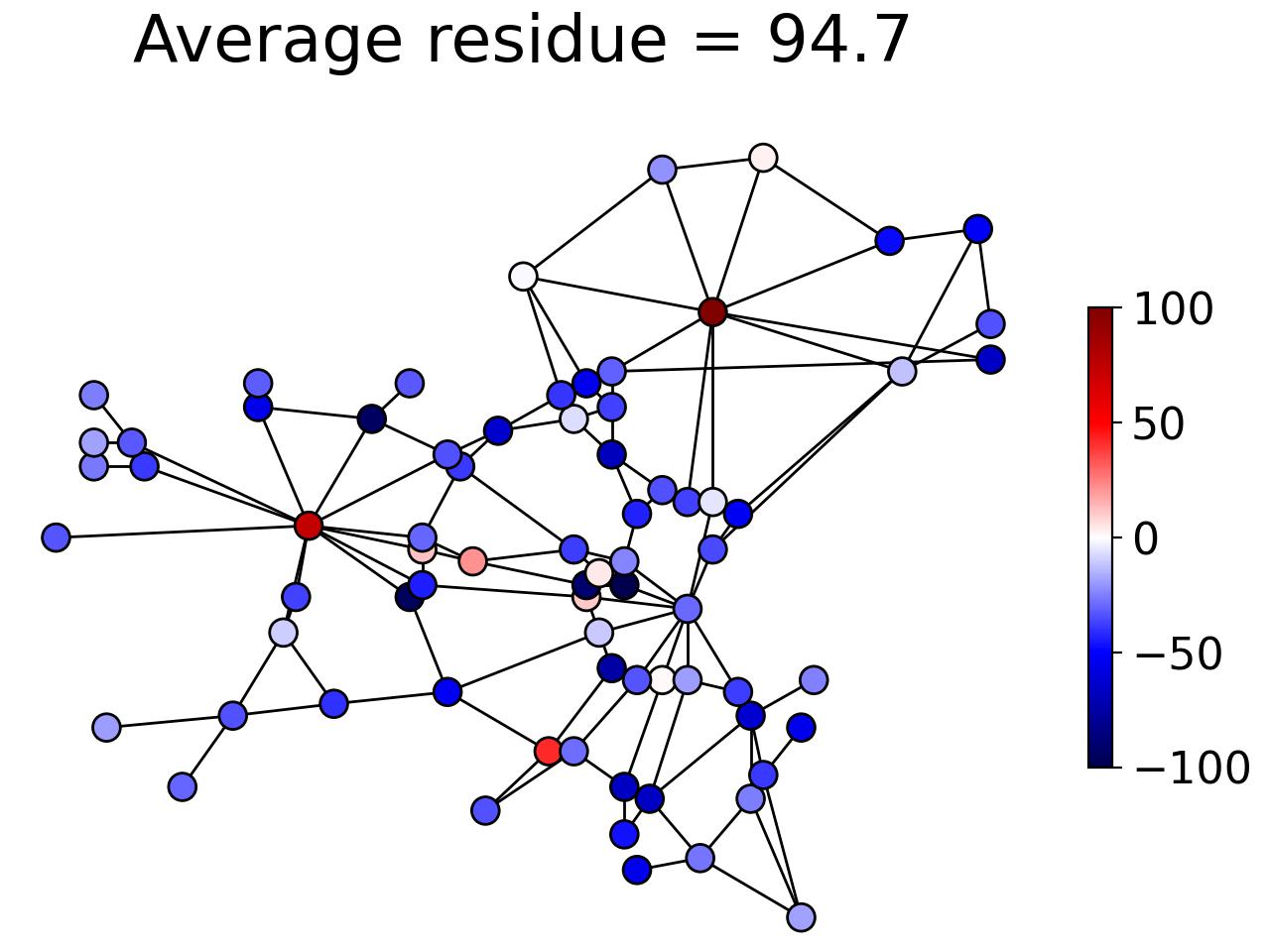}
    \caption{GCN}
    \label{fig:worst_residue_3}
\end{subfigure}
\hspace{5em}
\begin{subfigure}[b]{0.3\textwidth}
    \centering
    \includegraphics[width=\textwidth]{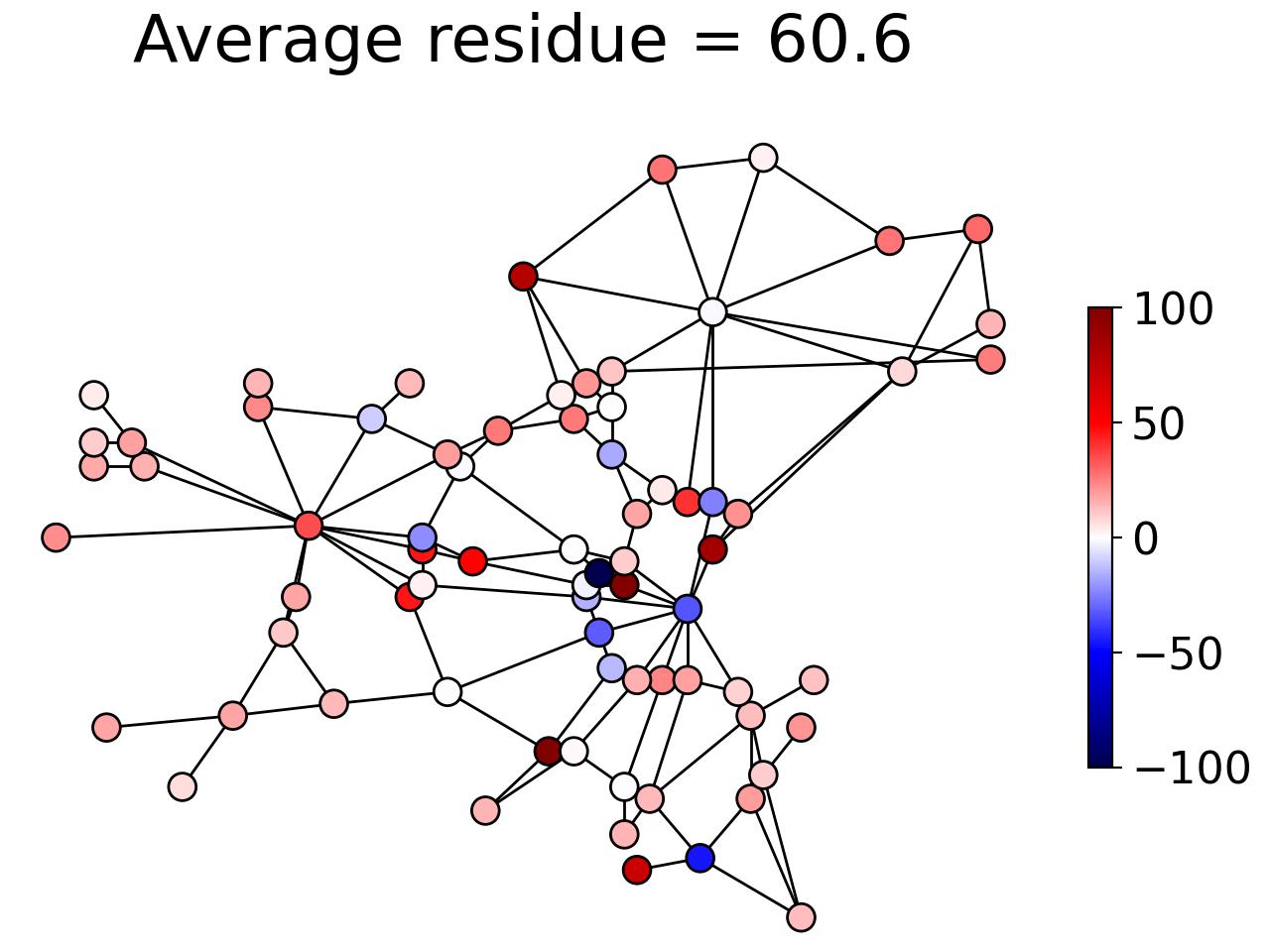}
    \caption{GraphSAGE}
    \label{fig:worst_residue_4}
\end{subfigure}
\caption{Comparison of node-level flow conservation residue in the East Massachusetts network. Four models are included: GAT, GCN, GraphSAGE, and ours}
\label{fig:residue}
\end{figure}

Additionally, Figure~\ref{fig:residue} visualizes the conservation residues at the node level. Specifically, we show the violation to the conservation low at each node in the EMA network when different models calculates the system optimal flows. As can be seen, our proposed  M-HetGAT model demonstrates superior in adheroing to the flow conservation, compared to its homogeneous counterparts, showing a 13\% improvement over the next best model. Moreover, the average normalized flow conservation residue relative to the total OD demand is 2.91\%. This suggests that the M-HetGAT model not only improves the accuracy of flow predictions but also ensures better compliance with the physical and network constraints.

\subsubsection{Out of distribution testing on modified network}
\label{sec:ood}

In this section, the training set consists of a transportation network where one or two links were randomly removed, whereas the testing set comprises networks with three links randomly removed. However, the training setup and hyperparameters remain identical to those in the previous experiments discussed in Section \ref{sec:setup}. The results, as shown in Table \ref{tab:urban_unseen}, indicate that our proposed M-HetGAT model consistently outperforms the baseline models including M-GAT, M-GCN, and M-GraphSAGE, across all networks, under both user equilibrium and system optimal scenarios. To explain further, the graph neural network identifies the missing links in the network, and the updated node features are processed through the intra-view and inter-view graph attention networks. Despite the network topologies are not seen in the training, our model still achieves accurate link flow predictions. This is especially notable compared to homogeneous GNN models, as our M-HetGAT model effectively captures the influence of OD pairs through virtual links. Consequently, it is more expressive and better equipped to model and learn the impact of topological alterations on flow distributions.

\begin{table}[htb!]
\renewcommand{\arraystretch}{1.15}
\centering
\caption{Out of distribution testing performance comparison of the proposed model with benchmark methods. Three networks are considered including Sioux Falls, East Massachusetts, and Anaheim. The mean absolute error, root mean square error is used to evaluate the prediction performance on the testing set.}
\label{tab:urban_unseen}
\resizebox{\textwidth}{!}{%
\begin{tabular}{ccccccc|cccc|cccc}
\hline
\multirow{3}{*}{Setting} & \multirow{3}{*}{Metric} & \multirow{3}{*}{Model} & \multicolumn{4}{c}{Sioux   Falls} & \multicolumn{4}{c}{EMA} & \multicolumn{4}{c}{Anaheim} \\ \cline{4-15} 
 &  &  & \multicolumn{2}{c}{Car} & \multicolumn{2}{c}{Truck} & \multicolumn{2}{c}{Car} & \multicolumn{2}{c}{Truck} & \multicolumn{2}{c}{Car} & \multicolumn{2}{c}{Truck} \\ \cline{4-15} 
 &  &  & MAE & RMSE & MAE & RMSE & MAE & RMSE & MAE & RMSE & MAE & RMSE & MAE & RMSE \\ \hline
\multirow{8}{*}{\begin{tabular}[c]{@{}c@{}}System\\ optimal\end{tabular}} & \multirow{4}{*}{Flow} & M-GAT & 8.28 & 11.65 & 7.20 & 9.60 & 0.76 & 1.20 & 0.67 & 1.03 & 2.88 & 4.04 & 0.59 & 0.84 \\
 &  & M-GCN & 10.05 & 14.14 & 9.53 & 12.90 & 1.23 & 2.13 & 1.24 & 2.02 & 2.82 & 3.92 & 0.63 & 0.94 \\
 &  & M-GraphSAGE & 8.03 & 10.89 & 6.14 & 8.26 & 0.78 & 1.29 & 0.80 & 1.31 & 2.87 & 4.06 & \textbf{0.38} & \textbf{0.57} \\
 &  & M-HetGAT & \textbf{7.44} & \textbf{10.59} & \textbf{5.51} & \textbf{7.57} & \textbf{0.42} & \textbf{0.88} & \textbf{0.34} & \textbf{0.76} & \textbf{2.76} & \textbf{3.76} & 0.46 & 0.64 \\ \cline{2-15} 
 & \multirow{4}{*}{Ratio} & M-GAT & 5.97\% & 8.55\% & 5.28\% & 7.28\% & 2.39\% & 3.57\% & 2.20\% & 3.36\% & 3.44\% & 4.76\% & 0.99\% & 1.39\% \\
 &  & M-GCN & 7.18\% & 10.00\% & 6.59\% & 8.75\% & 3.91\% & 6.19\% & 4.02\% & 6.16\% & 2.52\% & 3.74\% & 0.78\% & 1.17\% \\
 &  & M-GraphSAGE & 5.90\% & 8.36\% & 4.37\% & 6.01\% & 2.35\% & 3.64\% & 2.40\% & 3.70\% & 3.03\% & 4.31\% & 1.01\% & 1.29\% \\
 &  & M-HetGAT & \textbf{5.33\%} & \textbf{7.70\%} & \textbf{3.96\%} & \textbf{5.54\%} & \textbf{1.12\%} & \textbf{1.99\%} & \textbf{0.91\%} & \textbf{1.76\%} & \textbf{2.08\%} & \textbf{3.07\%} & \textbf{0.43\%} & \textbf{0.62\%} \\ \hline
\multirow{8}{*}{\begin{tabular}[c]{@{}c@{}}User\\ equilibrium\end{tabular}} & \multirow{4}{*}{Flow} & M-GAT & 6.36 & 8.44 & 5.40 & 7.52 & 1.16 & 1.94 & 1.00 & 1.57 & 4.05 & 5.54 & 0.90 & 1.26 \\
 &  & M-GCN & 7.53 & 11.30 & 6.79 & 10.11 & 1.59 & 2.44 & 1.39 & 2.21 & 3.31 & 4.73 & 0.63 & 0.96 \\
 &  & M-GraphSAGE & 6.57 & 8.70 & \textbf{5.36} & \textbf{7.00} & 0.85 & 1.44 & 0.84 & 1.37 & 3.00 & 4.33 & 0.52 & 0.81 \\
 &  & M-HetGAT & \textbf{6.04} & \textbf{8.51} & 5.58 & 7.81 & \textbf{0.39} & \textbf{0.86} & \textbf{0.40} & \textbf{0.91} & \textbf{2.96} & \textbf{4.21} & \textbf{0.45} & \textbf{0.67} \\ \cline{2-15} 
 & \multirow{4}{*}{Ratio} & M-GAT & 5.10\% & 7.39\% & 4.29\% & 6.54\% & 3.67\% & 5.94\% & 3.35\% & 5.79\% & 3.95\% & 6.12\% & 1.01\% & 1.63\% \\
 &  & M-GCN & 5.91\% & 9.47\% & 5.14\% & 8.24\% & 5.11\% & 7.65\% & 4.73\% & 7.36\% & 3.28\% & 5.40\% & 0.96\% & 1.46\% \\
 &  & M-GraphSAGE & 5.07\% & \textbf{7.19\%} & \textbf{4.02\%} & \textbf{5.66\%} & 2.66\% & 4.66\% & 2.78\% & 4.94\% & 3.00\% & 4.35\% & 0.89\% & 1.30\% \\
 &  & M-HetGAT & \textbf{4.83\%} & 7.39\% & 4.37\% & 6.58\% & \textbf{1.09\%} & \textbf{2.27\%} & \textbf{1.09\%} & \textbf{2.33\%} & \textbf{2.41\%} & \textbf{3.35\%} & \textbf{0.55\%} & \textbf{0.76\%} \\ \hline
\end{tabular}%
}
\end{table}

In addition to prediction accuracy, training time is a critical factor when assessing the efficiency and practicality of machine learning models. The overall computational time can be broken down into three key components: (1) the time spent solving traffic assignment problems, (2) the training time for the GNN model, and (3) the GNN's inference time. For every 1,000 graphs from the Sioux Falls, EMA, and Anaheim networks, the time required to solve multi-class vehicle TAP is 56.8, 575.2, and 2,612.5 minutes, respectively. In contrast, the training time of the M-HetGAT model for the same graphs is significantly lower, at 26.8, 28.9, and 59.7 minutes, respectively. Furthermore, the inference time for the M-HetGAT is particularly efficient, taking just 0.13, 0.15, and 0.31 minutes, respectively, for the same set of graphs. This breakdown of computational time highlights the overall efficiency of the proposed model in addressing traffic assignment problems.

\subsection{Ablation Study}
To evaluate the effectiveness of each component within the M-HetGAT framework, we conducted an ablation study by excluding various components of the model one at a time. In particular, we considered five cases: (1) ``w/o link feat", which is M-HetGAT without road link features, thereby omitting the consideration of link-specific variations; (2) ``w/o OD link", which is M-HetGAT without any OD link, forming a homogeneous graph with only physical links; and (3) ``w/o Intra view", which is M-HetGAT after removing intra-view message passing; (4) ``w/o Multi-head", which is M-HetGAT with only single head attention mechanism; (5) ``w/o conservation", which remove the flow conservation loss in the back propagation. To ensure a comprehensive evaluation, we conducted experiments with all variants across all three networks, under both system optimal scenarios and user equilibrium scenarios. Specifically, we compared the mean absolute error of the link utilization ratio among different configurations. This approach allowed us to identify the impact of each component on the model's overall performance.

\begin{figure}[htb!]
\centering
\includegraphics[width=0.95\textwidth]{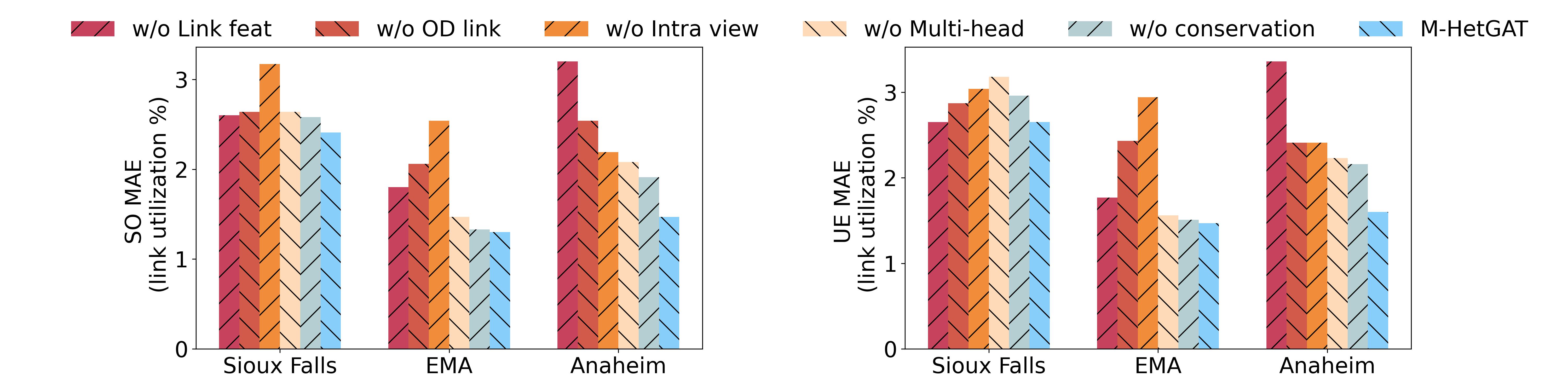}
\caption{Ablation study of different variants of M-HetGAT. Three variations of M-HetGAT are included: ``w/o link feat": model maintaining OD link but removing road link features; ``w/o OD link": model removing OD link; ``w/o Intra view": model removing intra-view message passing; ``w/o Multi-head": model removing multi-head attention mechanism; ``w/o conservation": model removing flow conservation in the loss function }
\label{fig:ablation}
\end{figure}

The comparative analysis, illustrated in Figure \ref{fig:ablation}, highlights a degradation in performance across all modified models compared to the complete model, emphasizing the critical contribution of each individual component. Notably, the inclusion of OD links proves to be particularly influential, as their removal results in significant increases in error for flow and link utilization predictions. This can be attributed to the fact that integrating OD links into the M-HetGAT model reduces the need for excessive embedding propagation through road links, thereby enhancing model efficiency.

Furthermore, removing link features from the OD links also leads to a performance decline, underscoring the importance of link features in accurately predicting traffic flow. Similarly, the removal of intra-view message passing and multi-head attention mechanisms from the models also results in decreased performance compared to the full model. Specifically, intra-view message passing, which is designed for multi-class traffic assignment, facilitates the adaptive representation of link features, thereby improving the model's ability to capture dynamic traffic flow distribution. Additionally, excluding the conservation loss from the overall loss function further degrades performance, indicating that the regularization term plays a crucial role in improving the model’s overall effectiveness.

\section{Conclusion and Discussion}
\label{sec:conslusion}

This study introduces a novel multi-view heterogeneous graph neural network (M-HetGAT) framework to tackle the computational challenges of multi-class traffic assignment problems in large-scale transportation networks. The framework employs a multi-view graph attention mechanism designed to accommodate different vehicle classes and integrates a conservation-based loss function, ensuring high accuracy in predicting link flows and utilization ratios while maintaining adherence to core traffic flow principles. Comprehensive experiments on urban road networks validate the model’s efficacy and generalizability, showcasing significant improvements in convergence speed and predictive accuracy over conventional graph neural network models. Moreover, the study highlights the model's ability to address real-world complexities, including varying road accessibility for diverse vehicle classes and dynamic network conditions, such as road closures and topology changes.

The proposed framework serves as a surrogate model capable of significantly accelerating complex optimization tasks in domains such as resource allocation and infrastructure asset management. While the current model is designed for static origin-destination demand data, future research could explore its extension to dynamic demand conditions and real-time traffic updates. Incorporating external influences such as weather, special events, and socio-economic factors may further broaden the model’s practical applicability.

\section{Acknowledgment}
This work was supported in part by the National Science Foundation under Grant CMMI-1752302.

\bibliographystyle{elsarticle-num-names} 
\bibliography{cas-refs}

\begin{thebibliography}{28}
\expandafter\ifx\csname natexlab\endcsname\relax\def\natexlab#1{#1}\fi
\providecommand{\url}[1]{\texttt{#1}}
\providecommand{\href}[2]{#2}
\providecommand{\path}[1]{#1}
\providecommand{\DOIprefix}{doi:}
\providecommand{\ArXivprefix}{arXiv:}
\providecommand{\URLprefix}{URL: }
\providecommand{\Pubmedprefix}{pmid:}
\providecommand{\doi}[1]{\href{http://dx.doi.org/#1}{\path{#1}}}
\providecommand{\Pubmed}[1]{\href{pmid:#1}{\path{#1}}}
\providecommand{\bibinfo}[2]{#2}
\ifx\xfnm\relax \def\xfnm[#1]{\unskip,\space#1}\fi
\bibitem[{Nie et~al.(2004)Nie, Zhang, and Lee}]{nie2004models}
\bibinfo{author}{Y.~Nie}, \bibinfo{author}{H.~Zhang}, \bibinfo{author}{D.-H.
  Lee},
\newblock \bibinfo{title}{Models and algorithms for the traffic assignment
  problem with link capacity constraints},
\newblock \bibinfo{journal}{Transportation Research Part B: Methodological}
  \bibinfo{volume}{38} (\bibinfo{year}{2004}) \bibinfo{pages}{285--312}.
\bibitem[{Seliverstov et~al.(2017)Seliverstov, Seliverstov, Malygin, Tarantsev,
  Shatalova, Lukomskaya, Tishchenko, and
  Elyashevich}]{seliverstov2017development}
\bibinfo{author}{Y.~A. Seliverstov}, \bibinfo{author}{S.~A. Seliverstov},
  \bibinfo{author}{I.~G. Malygin}, \bibinfo{author}{A.~A. Tarantsev},
  \bibinfo{author}{N.~V. Shatalova}, \bibinfo{author}{O.~Y. Lukomskaya},
  \bibinfo{author}{I.~P. Tishchenko}, \bibinfo{author}{A.~M. Elyashevich},
\newblock \bibinfo{title}{Development of management principles of urban traffic
  under conditions of information uncertainty},
\newblock in: \bibinfo{booktitle}{Creativity in Intelligent Technologies and
  Data Science: Second Conference, CIT\&DS 2017, Volgograd, Russia, September
  12-14, 2017, Proceedings 2}, \bibinfo{organization}{Springer},
  \bibinfo{year}{2017}, pp. \bibinfo{pages}{399--418}.
\bibitem[{Zou and Chen(2020)}]{zou2020resilience}
\bibinfo{author}{Q.~Zou}, \bibinfo{author}{S.~Chen},
\newblock \bibinfo{title}{Resilience modeling of interdependent
  traffic-electric power system subject to hurricanes},
\newblock \bibinfo{journal}{Journal of Infrastructure Systems}
  \bibinfo{volume}{26} (\bibinfo{year}{2020}) \bibinfo{pages}{04019034}.
\bibitem[{Liu and Meidani(2023)}]{liu2023optimizing}
\bibinfo{author}{T.~Liu}, \bibinfo{author}{H.~Meidani},
\newblock \bibinfo{title}{Optimizing seismic retrofit of bridges: Integrating
  efficient graph neural network surrogates and transportation equity},
\newblock in: \bibinfo{booktitle}{Proceedings of Cyber-Physical Systems and
  Internet of Things Week 2023}, \bibinfo{year}{2023}, pp.
  \bibinfo{pages}{367--372}.
\bibitem[{Lee et~al.(2003)Lee, Nie, and Chen}]{lee2003conjugate}
\bibinfo{author}{D.-H. Lee}, \bibinfo{author}{Y.~Nie},
  \bibinfo{author}{A.~Chen},
\newblock \bibinfo{title}{A conjugate gradient projection algorithm for the
  traffic assignment problem},
\newblock \bibinfo{journal}{Mathematical and computer modelling}
  \bibinfo{volume}{37} (\bibinfo{year}{2003}) \bibinfo{pages}{863--878}.
\bibitem[{Zhang et~al.(2023)Zhang, Liu, and D’Ariano}]{zhang2023bi}
\bibinfo{author}{Q.~Zhang}, \bibinfo{author}{S.~Q. Liu},
  \bibinfo{author}{A.~D’Ariano},
\newblock \bibinfo{title}{Bi-objective bi-level optimization for integrating
  lane-level closure and reversal in redesigning transportation networks},
\newblock \bibinfo{journal}{Operational Research} \bibinfo{volume}{23}
  (\bibinfo{year}{2023}) \bibinfo{pages}{23}.
\bibitem[{Fan et~al.(2023)Fan, Tang, Ye, Xiao, and Zhang}]{fan2023deep}
\bibinfo{author}{W.~Fan}, \bibinfo{author}{Z.~Tang}, \bibinfo{author}{P.~Ye},
  \bibinfo{author}{F.~Xiao}, \bibinfo{author}{J.~Zhang},
\newblock \bibinfo{title}{Deep learning-based dynamic traffic assignment with
  incomplete origin--destination data},
\newblock \bibinfo{journal}{Transportation Research Record}
  \bibinfo{volume}{2677} (\bibinfo{year}{2023}) \bibinfo{pages}{1340--1356}.
\bibitem[{Rahman and Hasan(2023)}]{rahman2023data}
\bibinfo{author}{R.~Rahman}, \bibinfo{author}{S.~Hasan},
\newblock \bibinfo{title}{Data-driven traffic assignment: A novel approach for
  learning traffic flow patterns using graph convolutional neural network},
\newblock \bibinfo{journal}{Data Science for Transportation}
  \bibinfo{volume}{5} (\bibinfo{year}{2023}) \bibinfo{pages}{11}.
\bibitem[{Liu and Meidani(2024)}]{liu2024end}
\bibinfo{author}{T.~Liu}, \bibinfo{author}{H.~Meidani},
\newblock \bibinfo{title}{End-to-end heterogeneous graph neural networks for
  traffic assignment},
\newblock \bibinfo{journal}{Transportation Research Part C: Emerging
  Technologies} \bibinfo{volume}{165} (\bibinfo{year}{2024})
  \bibinfo{pages}{104695}.
\bibitem[{Babazadeh et~al.(2020)Babazadeh, Javani, Gentile, and
  Florian}]{babazadeh2020reduced}
\bibinfo{author}{A.~Babazadeh}, \bibinfo{author}{B.~Javani},
  \bibinfo{author}{G.~Gentile}, \bibinfo{author}{M.~Florian},
\newblock \bibinfo{title}{Reduced gradient algorithm for user equilibrium
  traffic assignment problem},
\newblock \bibinfo{journal}{Transportmetrica A: Transport Science}
  \bibinfo{volume}{16} (\bibinfo{year}{2020}) \bibinfo{pages}{1111--1135}.
\bibitem[{Sun and Szeto(2021)}]{sun2021multi}
\bibinfo{author}{S.~Sun}, \bibinfo{author}{W.~Szeto},
\newblock \bibinfo{title}{Multi-class stochastic user equilibrium assignment
  model with ridesharing: Formulation and policy implications},
\newblock \bibinfo{journal}{Transportation Research Part A: Policy and
  Practice} \bibinfo{volume}{145} (\bibinfo{year}{2021})
  \bibinfo{pages}{203--227}.
\bibitem[{Xu et~al.(2024)Xu, Peng, Li, Chen, and Liu}]{xu2024range}
\bibinfo{author}{Z.~Xu}, \bibinfo{author}{Y.~Peng}, \bibinfo{author}{G.~Li},
  \bibinfo{author}{A.~Chen}, \bibinfo{author}{X.~Liu},
\newblock \bibinfo{title}{Range-constrained traffic assignment for electric
  vehicles under heterogeneous range anxiety},
\newblock \bibinfo{journal}{Transportation Research Part C: Emerging
  Technologies} \bibinfo{volume}{158} (\bibinfo{year}{2024})
  \bibinfo{pages}{104419}.
\bibitem[{Zhang et~al.(2018)Zhang, Yuan, Zeng, Li, and Wei}]{zhang2018missing}
\bibinfo{author}{Q.~Zhang}, \bibinfo{author}{Q.~Yuan},
  \bibinfo{author}{C.~Zeng}, \bibinfo{author}{X.~Li}, \bibinfo{author}{Y.~Wei},
\newblock \bibinfo{title}{Missing data reconstruction in remote sensing image
  with a unified spatial--temporal--spectral deep convolutional neural
  network},
\newblock \bibinfo{journal}{IEEE Transactions on Geoscience and Remote Sensing}
  \bibinfo{volume}{56} (\bibinfo{year}{2018}) \bibinfo{pages}{4274--4288}.
\bibitem[{Liu and Meidani(2024{\natexlab{a}})}]{liu2024graph}
\bibinfo{author}{T.~Liu}, \bibinfo{author}{H.~Meidani},
\newblock \bibinfo{title}{Graph neural network surrogate for seismic
  reliability analysis of highway bridge systems},
\newblock \bibinfo{journal}{Journal of Infrastructure Systems}
  \bibinfo{volume}{30} (\bibinfo{year}{2024}{\natexlab{a}})
  \bibinfo{pages}{05024004}.
\bibitem[{Liu and Meidani(2024{\natexlab{b}})}]{liu2024heterogeneous}
\bibinfo{author}{T.~Liu}, \bibinfo{author}{H.~Meidani},
\newblock \bibinfo{title}{Heterogeneous graph sequence neural networks for
  dynamic traffic assignment},
\newblock \bibinfo{journal}{arXiv preprint arXiv:2408.04131}
  (\bibinfo{year}{2024}{\natexlab{b}}).
\bibitem[{Beckmann et~al.(1956)Beckmann, McGuire, and
  Winsten}]{beckmann1956studies}
\bibinfo{author}{M.~Beckmann}, \bibinfo{author}{C.~B. McGuire},
  \bibinfo{author}{C.~B. Winsten}, \bibinfo{title}{Studies in the Economics of
  Transportation}, \bibinfo{type}{Technical Report}, \bibinfo{year}{1956}.
\bibitem[{Liu et~al.(2020)Liu, Lyu, Liu, and Liu}]{liu2020automatic}
\bibinfo{author}{Y.~Liu}, \bibinfo{author}{C.~Lyu}, \bibinfo{author}{X.~Liu},
  \bibinfo{author}{Z.~Liu},
\newblock \bibinfo{title}{Automatic feature engineering for bus passenger flow
  prediction based on modular convolutional neural network},
\newblock \bibinfo{journal}{IEEE Transactions on Intelligent Transportation
  Systems} \bibinfo{volume}{22} (\bibinfo{year}{2020})
  \bibinfo{pages}{2349--2358}.
\bibitem[{Hamilton et~al.(2017)Hamilton, Ying, and
  Leskovec}]{hamilton2017inductive}
\bibinfo{author}{W.~Hamilton}, \bibinfo{author}{Z.~Ying},
  \bibinfo{author}{J.~Leskovec},
\newblock \bibinfo{title}{Inductive representation learning on large graphs},
\newblock \bibinfo{journal}{Advances in neural information processing systems}
  \bibinfo{volume}{30} (\bibinfo{year}{2017}).
\bibitem[{Veli{\v{c}}kovi{\'c} et~al.(2017)Veli{\v{c}}kovi{\'c}, Cucurull,
  Casanova, Romero, Lio, and Bengio}]{velivckovic2017graph}
\bibinfo{author}{P.~Veli{\v{c}}kovi{\'c}}, \bibinfo{author}{G.~Cucurull},
  \bibinfo{author}{A.~Casanova}, \bibinfo{author}{A.~Romero},
  \bibinfo{author}{P.~Lio}, \bibinfo{author}{Y.~Bengio},
\newblock \bibinfo{title}{Graph attention networks},
\newblock \bibinfo{journal}{arXiv preprint arXiv:1710.10903}
  (\bibinfo{year}{2017}).
\bibitem[{Wang et~al.(2019)Wang, Ji, Shi, Wang, Ye, Cui, and
  Yu}]{wang2019heterogeneous}
\bibinfo{author}{X.~Wang}, \bibinfo{author}{H.~Ji}, \bibinfo{author}{C.~Shi},
  \bibinfo{author}{B.~Wang}, \bibinfo{author}{Y.~Ye}, \bibinfo{author}{P.~Cui},
  \bibinfo{author}{P.~S. Yu},
\newblock \bibinfo{title}{Heterogeneous graph attention network},
\newblock in: \bibinfo{booktitle}{The world wide web conference},
  \bibinfo{year}{2019}, pp. \bibinfo{pages}{2022--2032}.
\bibitem[{Fu et~al.(2020)Fu, Zhang, Meng, and King}]{fu2020magnn}
\bibinfo{author}{X.~Fu}, \bibinfo{author}{J.~Zhang}, \bibinfo{author}{Z.~Meng},
  \bibinfo{author}{I.~King},
\newblock \bibinfo{title}{Magnn: Metapath aggregated graph neural network for
  heterogeneous graph embedding},
\newblock in: \bibinfo{booktitle}{Proceedings of The Web Conference 2020},
  \bibinfo{year}{2020}, pp. \bibinfo{pages}{2331--2341}.
\bibitem[{Bar-Gera et~al.(2023)Bar-Gera, Stabler, and
  Sall}]{bar2021transportation}
\bibinfo{author}{H.~Bar-Gera}, \bibinfo{author}{B.~Stabler},
  \bibinfo{author}{E.~Sall},
\newblock \bibinfo{title}{Transportation networks for research core team},
\newblock \bibinfo{journal}{Transportation Network Test Problems. Available
  online: https://github.com/bstabler/TransportationNetworks (accessed on May
  14 2023)}  (\bibinfo{year}{2023}).
\bibitem[{Pulugurtha and Jain(2022)}]{pulugurtha2022passenger}
\bibinfo{author}{S.~S. Pulugurtha}, \bibinfo{author}{R.~N. Jain},
\newblock \bibinfo{title}{Passenger car equivalent travel time of a truck},
\newblock \bibinfo{journal}{Multimodal Transportation} \bibinfo{volume}{1}
  (\bibinfo{year}{2022}) \bibinfo{pages}{100031}.
\bibitem[{Fukushima(1984)}]{fukushima1984modified}
\bibinfo{author}{M.~Fukushima},
\newblock \bibinfo{title}{A modified frank-wolfe algorithm for solving the
  traffic assignment problem},
\newblock \bibinfo{journal}{Transportation Research Part B: Methodological}
  \bibinfo{volume}{18} (\bibinfo{year}{1984}) \bibinfo{pages}{169--177}.
\bibitem[{Paszke et~al.(2019)Paszke, Gross, Massa, Lerer, Bradbury, Chanan,
  Killeen, Lin, Gimelshein, Antiga et~al.}]{paszke2019pytorch}
\bibinfo{author}{A.~Paszke}, \bibinfo{author}{S.~Gross},
  \bibinfo{author}{F.~Massa}, \bibinfo{author}{A.~Lerer},
  \bibinfo{author}{J.~Bradbury}, \bibinfo{author}{G.~Chanan},
  \bibinfo{author}{T.~Killeen}, \bibinfo{author}{Z.~Lin},
  \bibinfo{author}{N.~Gimelshein}, \bibinfo{author}{L.~Antiga}, et~al.,
\newblock \bibinfo{title}{Pytorch: An imperative style, high-performance deep
  learning library},
\newblock \bibinfo{journal}{Advances in neural information processing systems}
  \bibinfo{volume}{32} (\bibinfo{year}{2019}).
\bibitem[{Wang et~al.(2019)Wang, Zheng, Ye, Gan, Li, Song, Zhou, Ma, Yu, Gai
  et~al.}]{wang2019deep}
\bibinfo{author}{M.~Wang}, \bibinfo{author}{D.~Zheng}, \bibinfo{author}{Z.~Ye},
  \bibinfo{author}{Q.~Gan}, \bibinfo{author}{M.~Li}, \bibinfo{author}{X.~Song},
  \bibinfo{author}{J.~Zhou}, \bibinfo{author}{C.~Ma}, \bibinfo{author}{L.~Yu},
  \bibinfo{author}{Y.~Gai}, et~al.,
\newblock \bibinfo{title}{Deep graph library: A graph-centric,
  highly-performant package for graph neural networks},
\newblock \bibinfo{journal}{arXiv preprint arXiv:1909.01315}
  (\bibinfo{year}{2019}).
\bibitem[{Liu and Meidani(2023)}]{liu2023physics1}
\bibinfo{author}{T.~Liu}, \bibinfo{author}{H.~Meidani},
\newblock \bibinfo{title}{Physics-informed neural network for nonlinear
  structural system identification},
\newblock in: \bibinfo{booktitle}{14th International Workshop on Structural
  Health Monitoring: Designing SHM for Sustainability, Maintainability, and
  Reliability, IWSHM 2023}, \bibinfo{organization}{DEStech Publications},
  \bibinfo{year}{2023}, pp. \bibinfo{pages}{3001--3011}.
\bibitem[{Rodrigue(2020)}]{rodrigue2020geography}
\bibinfo{author}{J.-P. Rodrigue}, \bibinfo{title}{The geography of transport
  systems}, \bibinfo{publisher}{Routledge}, \bibinfo{year}{2020}.

\end{thebibliography}

\end{document}